\documentclass[journal,11pt,onecolumn,draftclsnofoot,]{IEEEtran}

\ifCLASSINFOpdf
  \usepackage[pdftex]{graphicx}

  \graphicspath{{./figures}}

\else

  \usepackage[dvips]{graphicx}

  \graphicspath{{./figures}}
\fi

\usepackage{gensymb}
\usepackage{color}

\usepackage{setspace}
\usepackage{adjustbox}

\usepackage{booktabs}

\newcommand{\ra}[1]{\renewcommand{\arraystretch}{#1}}

\begin{document}

\title{A practical guide to CNNs and Fisher Vectors \\
for image instance retrieval}

% Sinlge column affiliation block
\author{
\IEEEauthorblockN{Vijay Chandrasekhar{$^{*,1}$}, Jie Lin{$^{*,1}$}, Olivier Mor\`{e}re{$^{*,1,2}$}\\
Hanlin Goh{$^{1}$}, Antoine Veillard{$^{2}$}}\\
\and
\IEEEauthorblockA{{$^{1}$}Institute for Infocomm Research (A*STAR), 1 Fusionolopis Way, \#21-01, 138632, Singapore}\\
\IEEEauthorblockA{{$^{2}$}Universit\'e Pierre et Marie Curie, 4 place Jussieu, 75252, Paris, France}
\thanks{{$^{*}$} V. Chandrasekhar, J. Lin and O. Mor\`ere contributed equally to this work.}% <-this % stops a space
}

%\markboth{
%Jan 2015
%}
%{Shell \MakeLowercase{\textit{et al.}}: Bare Demo of IEEEtran.cls for Journals}

\maketitle

\vspace*{-4pc}
\begin{abstract}

%NEW
The computation of good image descriptors is key to the instance retrieval problem and has been the object of much recent interest from the multimedia research community.
With deep learning becoming the dominant approach in computer vision, the use of representations extracted from Convolutional Neural Nets (CNNs) is quickly gaining ground on Fisher Vectors (FVs) as favoured state-of-the-art global image descriptors for image instance retrieval.
While the good performance of CNNs for image classification are unambiguously recognised, which of the two has the upper hand in the image retrieval context is not entirely clear yet.

In this work, we propose a comprehensive study that systematically evaluates FVs and CNNs for image retrieval.
The first part compares the performances of FVs and CNNs on multiple publicly available data sets.
We investigate a number of details specific to each method.
For FVs, we compare sparse descriptors based on interest point detectors with dense single-scale and multi-scale variants.
For CNNs, we focus on understanding the impact of depth, architecture and training data on retrieval results.
Our study shows that no descriptor is systematically better than the other and that performance gains can usually be obtained by using both types together.
The second part of the study focuses on the impact of geometrical transformations such as rotations and scale changes.
FVs based on interest point detectors are intrinsically resilient to such transformations while CNNs do not have a built-in mechanism to ensure such invariance.
We show that performance of CNNs can quickly degrade in presence of rotations while they are far less affected by changes in scale.
We then propose a number of ways to incorporate the required invariances in the CNN pipeline.

Overall, our work is intended as a reference guide offering practically useful and simply implementable guidelines to anyone looking for state-of-the-art global descriptors best suited to their specific image instance retrieval problem.

\end{abstract}

\vspace*{-.5pc}
\begin{IEEEkeywords}
\vspace*{-.2pc}
convolutional neural networks, Fisher vectors, image instance retrieval.
\end{IEEEkeywords}

\IEEEpeerreviewmaketitle

\begin{table}[p]
\caption{Summary of Experimental Results and Key Findings}
\label{tab:summary_results}
\centering
\ra{1.2} 
{\footnotesize \singlespacing
\begin{adjustbox}{max width=\textwidth,center}
\begin{tabular}{@{}p{2.4in}p{4.3in}@{}}
\toprule

{\sc Questions} & {\sc Observations and Recommendations}\\

\midrule

\multicolumn{2}{@{}l@{}}{\bf Best practices for CNN descriptors}\\

\midrule

\emph{Best single crop strategy?} &
The largest possible center crop (discarding parts of the image but preserving aspect ratio) or the entire image (preserving the entire image but ignoring aspect ratio) work comparably, both outperforming padding (preserving both).
\\

\emph{Best performing layer?} & 
The first fully connected layer is a good all-round choice on all the tested models.\\

\emph{Do deeper networks help?} & 
Only if the training and test data are similar. Else, extra-depth can hurt performance. \\

\emph{How much does training data matter?} &
Training data has significant impact on performance. Results also suggest that deeper layers are more domain specific.\\

\midrule

\multicolumn{2}{@{}l@{}}{\bf Best practices for FV interest points}\\

\midrule

\emph{Dense or sparse interest points?} &
It depends on the dataset.
If scale and rotation invariance are not required, and the data are highly textured, dense sampling outperforms DoG interest points. \\

\emph{Single-scale or multi-scale interest points?} &
Multi-scale interest points always improve performance.
\\

\midrule

\multicolumn{2}{@{}l@{}}{\bf CNN versus FV}\\

\midrule

\emph{How do state-of-the-art CNN and FV results compare on standard benchmarks?} &
It depends on the characteristics of the data set. \\

\emph{Does combining FV and CNN improve performance?} & 
Yes, combining FV with state-of-the-art CNN descriptors can improve retrieval performance often by a significant margin.\\

\midrule

\multicolumn{2}{@{}l@{}}{\bf Invariance to rotations}\\

\midrule

\emph{How invariant are CNN features to rotation?} & 
CNN features exhibit very limited invariance to rotation, performance drops rapidly as query rotation angle is varied.\\

\emph{Are CNNs or FVs more invariant to rotation?} & 
FV based on DoG interest points are robust to rotation changes, as would be expected.
CNN descriptors are more robust to rotation changes than FV based on dense sampling. \\

\emph{How do we gain rotation invariance for CNN features?} &
Max-pooling across rotated versions of database images works well, at the loss of some discriminativeness when query and database images are aligned.
However, the same max-pooling approach is not effective on dense FVs. \\

\emph{Are deeper CNN layers more invariant to rotation?} &
The fully connected layers exhibit similar invariance properties to rotation.
Visual features (\emph{pool5}) are slightly more robust to small rotation angles but significantly less robust to larger angles.\\

\midrule

\multicolumn{2}{@{}l@{}}{\bf Invariance to scale changes}\\

\midrule

\emph{How scale-invariant are CNN features?} & 
CNN descriptors are robust to scale change and work well even for small query scales.\\

\emph{Are CNNs or FVs more scale-invariant?} & 
CNN descriptors are more robust to scale changes than any FV.
All FV variants experience a much sharper drop in performance as query scale is decreased compared to CNN features. \\

\emph{How do we gain scale invariance for CNN features?} &
Similar to rotation invariance, max-pooling across scaled versions of database images works well for gaining scale invariance, at the cost of some discriminativeness. \\

\emph{Are deeper CNN layers more scale-invariant?} &
Visual features (\emph{pool5}) are more scale-invariant than the deeper fully connected layers. \\

\bottomrule
\end{tabular}
\end{adjustbox}
}
\end{table}

\section{Introduction}
\label{sec:intro}

%NEW
Image instance retrieval is the discovery of images from a database representing the same object or scene as the one depicted in a query image.
State-of-the-art image instance retrieval pipelines consist of two major blocks: first, a subset of images similar to the query are retrieved from the database, next, geometric consistency checks are applied to select the relevant images from the subset with high precision. 
The first step is based on the comparison of {\it global image descriptors}: high-dimensional vectors with up to tens of thousands of dimensions representing the image contents.
Better global descriptors are key to improving retrieval performance and has been the object of much recent interest from the multimedia research community with work on specific applications such as digital documents~\cite{tmm02}, mobile visual search~\cite{tmm02,tmm05}, distributed large scale search~\cite{tmm03} and compact descriptors for fast real-world applications~\cite{tmm01,tmm04}.

A popular global descriptor which achieves high performance is the Fisher Vector (FV)~\cite{Perronnin_CVPR_10}.
The FV is obtained by quantizing the set of local feature descriptors with a small codebook of 64-512 centroids, and aggregating first and second order residual statistics for features quantized to each centroid. 
The residual statistics from each centroid are concatenated together to obtain the high-dimensional global descriptor representation, typically 8192 to 65536 dimensions.
The performance increases as the dimensionality of the global descriptor increases, as shown in~\cite{Perronnin_CVPR_10}. 
FVs can be aggregated on descriptors extracted densely in the image~\cite{fishertheoryandpractice}, or around interest points like Difference-of-Gaussian (DoG) interest points~\cite{Lowe04}.
The former is popular for image classification tasks, while the latter is used in image retrieval as the DoG interest points provide invariance to scale and rotation.

As opposed to the carefully hand-crafted FVs, deep learning has achieved remarkable performance for large scale image classification~\cite{AlexNet,VeryDeepNeuralNets}. 
Deep learning has also achieved state-of-the-art results in many other visual tasks such as face recognition~\cite{deepface,deepid}, pedestrian detection~\cite{deeppedestrian} and pose estimation~\cite{deeppose}. 
In their recent work,  Babenko et al.~\cite{Yandex} propose using representations extracted from Convolutional Neural Nets (CNN) as a global descriptor for image retrieval, and show promising initial results for the approach.
In our work, we show how stacked Restricted Boltzmann Machines (RBM) and supervised fine-tuning can be used for generating extremely compact hashes from global descriptors obtained from CNNs for large scale image-retrieval~\cite{CompactGlobal}.

While deep learning has unquestionably become the dominant approach for image classification, the case for image retrieval has yet to be clearly settled.
The two types of descriptors being radically different in nature, one can expect them to behave very differently based on specific aspects of the problem.
On one hand, CNNs seem to obtain good retrieval results with more compact starting representation but many factors related to the network architecture or the training may come into play.
On the other hand, FVs may be more robust to training data and more invariant to certain geometrical transformations of the images.
In fact, some of the best reported instance retrieval performances are still based on hand-crafted features such as FVs~\cite{confused}. 

In this work, we perform a thorough investigation of approaches based on FVs and CNNs on multiple publicly available datasets and analyse the pros and cons of each.
The first part of the study determines best practices for FVs and CNNs on details specific to each of the approach.
For FVs, we investigate the effects of spare SIFT based on interest point detectors versus dense SIFT (single-scale and multi-scale).
For CNN descriptors, we specifically study the impacts of image cropping strategies, layer extracted from the CNN, network depth, and training data.
Next, we investigate how each type of descriptors performs compared to the other and if a combination of both types of descriptors can improve results over to the best FVs and CNN descriptors.
The final part of our work is dedicated to the impact of geometrical transformations such as rotations and scale changes.
Unlike FVs based on interest point detectors, CNNs do not have a built-in mechanism to ensure resilience to such transformations.
Hence it is necessary to understand how much CNN descriptors are affected by them.
We also propose a number of ways to incorporate transformation invariance in the CNN pipeline.

Our work provides a set of straightforward practical guidelines, some valid in general and some problem dependent,
one should follow to get the global image descriptors best suited to their specific image instance retrieval task.

\section{Related Work}
\label{sec:related}

There has been extensive work on the FV and its variants since it was first proposed for instance retrieval.
Several improvements to the baseline FV~\cite{Perronnin_CVPR_10} have been proposed in recent literature, including the Residual Enhanced Visual Vector~\cite{REVV1} and the Rate-adaptive Compact Fisher Codes (RCFC)~\cite{SFCV}.
Recent improvements also include better aggregation schemes~\cite{TEDA}, and better matching kernels~\cite{confused}.
State-of-the-art results using FVs are based on aggregating statistics around interest points like Difference-of-Gaussian~\cite{Lowe04} or Hessian-affine interest points~\cite{HessianSIFTCode}.

CNNs are now considered to be the mainstream approach for large-scale image classification. 
ImageNet 2014 submissions are all based on CNNs.
After the winning submission of Krizhevsky et al. in the ImageNet 2012 challenge~\cite{AlexNet}, CNN began to be applied to the instance retrieval problem as well.
There is comparatively less work on CNN-based descriptors for instance retrieval compared to large-scale image classification.
Razavian et al.~\cite{CNNOffTheShelf} evaluate the performance of CNN model of~\cite{AlexNet} on a wide range of tasks including instance retrieval, and show initial promising results.
Babenko et al.~\cite{Yandex} show that fine-tuning a pre-trained CNN with domain specific data (objects, scenes, etc) can improve retrieval performance on relevant data sets.
The authors also show that the CNN representations can be compressed more effectively than their Fisher counterparts for large-scale instance retrieval.
In~\cite{CompactGlobal}, we show how sparse high-dimensional CNN representations can be hashed to very compact representations (64-1024 bits) for large scale image retrieval with little loss in matching performance.

While the papers above show initial results, the CNN architecture and features from~\cite{AlexNet} are used as a black-box for the retrieval task.
There is no systematic study of how the CNN architecture and training data affect retrieval performance.
Also, unlike interest points which provide scale and rotation invariance to the FV pipeline, CNN representations used in image-classification are obtained by densely sampling a resized canonical image.
CNN features do not provide explicit rotation and scale invariance, which are often key to instance retrieval tasks.
Desired levels of scale and rotation invariance for CNN features can nevertheless be indirectly achieved from the max-pooling operations in the pipeline, the diversity of the training data which typically contains objects at varying scales and orientations, and data augmentation during the training phase where data can be preprocessed and input to the CNN at different scales and orientations.

\begin{figure}[ht]
\centering
\includegraphics[width=0.7\textwidth]{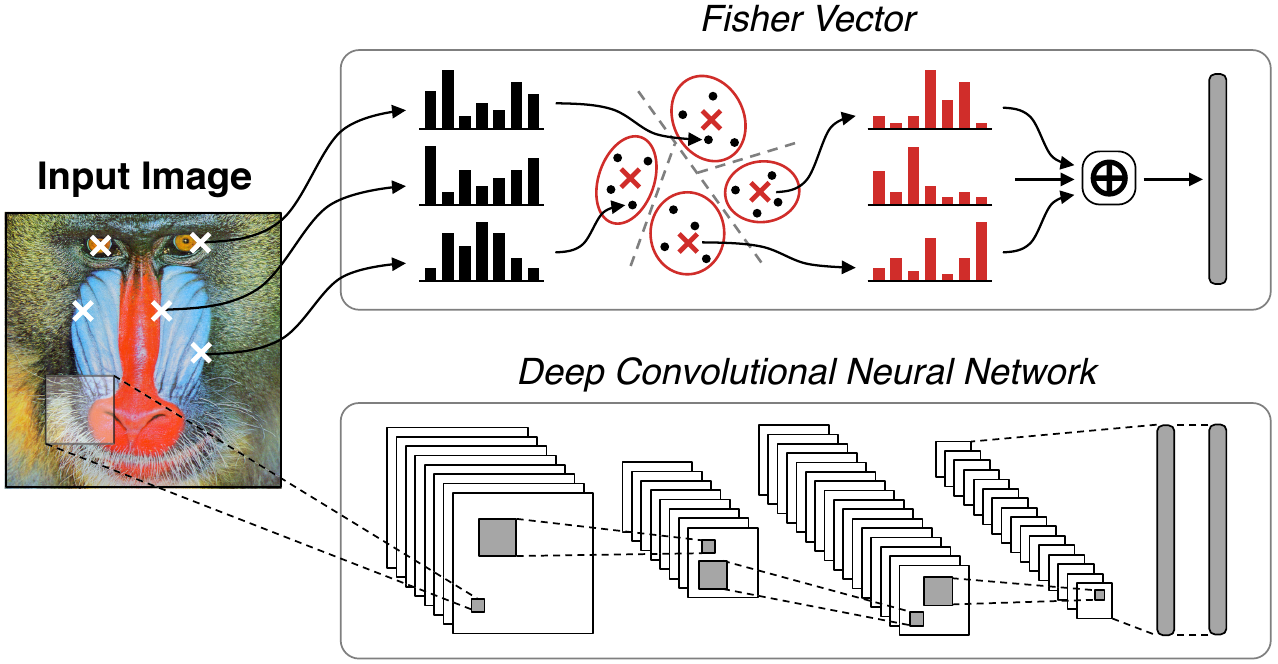}
\caption{FV and CNN based pipelines for the computation of global image descriptors.}
\label{fig:cnn_fv_pipelines}
\end{figure}

In this work, we provide a systematic and thorough evaluation of FV and CNN pipelines (see Figure~\ref{fig:cnn_fv_pipelines}) for instance retrieval. 
We run extensive experiments on 4 popular data sets: {\it Holidays}~\cite{Jegou08}, {\it UKBench}~\cite{Nister06}, {\it Oxford buildings}~\cite{Philbin07} and {\it Stanford Mobile Visual Search}~\cite{SVMSDataSet} to study how well CNN-based approaches generalize compared with FVs.
Our CNN experiments in this work are based on publicly available CNN models in Caffe~\cite{Caffe} and can be fully reproduced, unlike CNN models trained by Google, Baidu, Microsoft and Yandex in~\cite{GoogleCNN,BaiduCNN,Yandex,MicrosoftCNN}.

\section{Contributions}
\label{sec:contributions}

The main contributions of our work are summarized as follows:

\begin{itemize}

\item We provide a comprehensive and systematic evaluation of FVs and CNN descriptors for instance retrieval.
Our results based on many standard, publicly available dataset and various pre-trained state-of-the-art models for image classification are fully reproducible.

\item We identify the best practices for the use of each type of descriptors through a set of dedicated experiments.
For CNNs, we investigate the impacts of the image cropping strategy, the network depth, the layer selected as descriptors, and the training data.
For FVs, we study how densely sampled SIFT single-scale and multi-scale descriptors compare with against sparse interest point detectors.

\item We compare the best performing FVs and CNN descriptors from our study to various reported state-of-the-art results on the various datasets.
We also investigate if a mixture of FVs and CNN descriptors is able to further improve results.

\item Unlike FVs based on interest point detectors, CNNs do not have a built-in mechanism to ensure robustness to transformations such as rotations or scale changes.
We therefore conduct a set of experiments to compare the performance and robustness of the two types of descriptors when affected by rotations and scale changes.
We also propose a number of ways the descriptors could be made more invariant to those transformations.

\item The key findings from our study are summarized in Table~\ref{tab:summary_results}
intended as a quick reference guide for practical guidelines on the use of FVs and CNNs for image retrieval.
The guidelines are sometimes general but often dependent on specific characteristics of the problem which have been properly identified in this study.

\end{itemize}

\section{Evaluation Framework}
\label{sec:eval_framework}

\subsection{Data Sets}

We evaluate the performances of the descriptors against four popular data sets: {\it Holidays}, {\it Oxford buildings (Oxbuild)}, {\it UKBench} and {\it Graphics}.
The four datasets are chosen for the diversity of data they provide: {\it UKBench} and {\it Graphics} are object-centric featuring close-up shots of objects in indoor environments. 
{\it Holidays} and {\it Oxbuild} are scene-centric datasets consisting primarily of outdoor buildings and scenes.

{\textbf{INRIA Holidays.}
The INRIA Holidays dataset~\cite{Jegou08} consist of personal holiday pictures. 
The dataset includes a large variety of outdoor scene types: natural, man-made, water and fire effects. 
There are 500 queries and 991 database images.
Variations in lighting conditions are rare in this data set as the pictures from the same location are taken at the same time.

\textbf{Oxford Buildings.}
The Oxford Buildings Dataset~\cite{Philbin07} consists of 5062 images collected from Flickr representing landmark buildings in Oxford. 
The collection has been manually annotated to generate a comprehensive ground truth for 11 different landmarks, each represented by 5 possible queries. 
Note that the set contains 55 queries only. 

{\textbf{UKBench.}}
The University of Kentucky (UKY) data set~\cite{Nister06} consists of 2550 groups of common objects.
There are 4 images representing each.
Only the object of interest present in each image.
Thus, there is no foreground or background clutter within this data set.
All 10200 images are used as queries.

{\textbf{Graphics.}}
The Graphics data set is part of the Stanford Mobile Visual Search data set~\cite{SVMSDataSet}, which notably was used in the MPEG standard: Compact Descriptors for Visual Search (CDVS)~\cite{MPEGDataset2}.
The data set contains different categories of objects like CDs, DVDs, books, software products, business cards, etc.
For product categories (CDs, DVDs and books), at least one of the references is a clean version of the product obtained from the product website.
The query images include foreground and background clutter that would be considered typical in real-world scenarii, e.g., a picture of a CD might contain other CDs in the background.
This data set distinguishes from the other ones as it contains images of rigid objects captured under widely varying lighting conditions, perspective distortion, foreground and background clutter. 
Query images are taken with heterogeneous phone cameras. 
Each query has two relevant images.
There are 500 unique objects, 1500 queries, and 1000 database images.

\subsection{Fisher Vectors}

FVs are a concatenation of first and second order statistics of a set of feature descriptors quantized with a small codebook.
We resize all images (maintaining aspect ratio) so that the larger dimension of the image is equal to 640 pixels prior to FV extraction.
We use the implementation of FVs from the open source library VLFeat~\cite{VidaldiSIFT2}.
SIFT detectors and descriptors are also chosen from the same library.
The three different types of SIFT descriptors used to generate the FVs are Difference of Gaussians (DoG) SIFT, Dense Single-scale SIFT and Dense Multi-scale SIFT.

\begin{itemize}

\item \textbf{DoG SIFT}. 
We detect interest points in the DoG scale space, followed by 128-dimensional SIFT descriptors extracted from scaled and oriented patches centered on interest points.
Default peak and edge thresholds (0 and 10) are employed to filter out low contrast patches or patches close to the edge of the image.
Since the DoG detector extracts scale and rotation invariant interest points, it has been widely applied for the task of instance retrieval.
It is important to note that we do not use any feature selection algorithm to select a subset of ``good'' features - an approach that can result in a significant improvement in performance on the {\it Graphics} data set~\cite{FeatureSelection}.

\item \textbf{Dense Single-scale SIFT}. 
We extract SIFT descriptors from densely sampled patches (every 4 pixels) with fixed scale and upright orientation.
The patch size used for the extraction is $m \times s$ where $s$ is the scale parameter and $m$ is the magnification parameter.
We choose the default magnification parameter $m=6$ across all dense SIFT descriptors.
$s=4$ is chosen for single-scale SIFT.
Dense SIFT is faster to compute than DoG SIFT as the expensive interest point detection step is avoided - however, this comes at the cost of lower scale and rotation invariance.
Note that dense SIFT is mostly popular for image classification tasks.

\item \textbf{Dense Multi-Scale SIFT}.
We apply dense SIFT extraction at multiple resolutions ($s=\{4, 8, 12, 16\}$).
This is aimed at gaining some degree of scale invariance.

\end{itemize}

Closely following \cite{PQFisher}\cite{Perronnin_CVPR_10}, we apply dimensionality reduction on SIFT descriptors from 128 to 64 using PCA, and train a Gaussian Mixture Model (GMM) with 256 centroids.
Both first order (gradients w.r.t. mean) and second order (gradients w.r.t. variance) statistics are encoded to form the FV,
resulting in a $64 \times 256 \times 2 = 32768$-dimensional vector representation for each image.
Finally, we apply power law normalization to each component ($\alpha = 0.5$), followed by $L_2$ normalization to obtain the final normalized FV representation \cite{Perronnin_CVPR_10}.
Each dimension of the FV is stored as a floating point number.
No compression is applied.
We refer to the three FV as FVDoG (FV computed on DoG points), FVDS (FV computed densely at a single scale) and FVDM (FV computed densely at multiple scales) from here on. 

\subsection{Convolutional Neural Net features}

In this work, we consider four different pre-trained CNN models for the instance retrieval problem: 
\begin{itemize}

\item {\textbf {\it OxfordNet}}~\cite{Simonyan2014}: the best performing single network from the Oxford VGG team at ImageNet 2014.

\item {\textbf {\it AlexNet}}~\cite{AlexNet}: the model referenced as ``BVLC reference caffenet'' in the Caffe framework \cite{Caffe}.  
This model was the winning ImageNet submission of 2012.
This network closely mimics the original {\it AlexNet} model of~\cite{AlexNet}. 

\item {\textbf {\it PlacesNet}}~\cite{Zhou2014}: a state-of-the-art model for scene image classification providing highest accuracy on the SUN397 dataset~\cite{xiao2010sun}.

\item {\textbf {\it HybridNet}}~\cite{Zhou2014}: another model for both object and scene images classification, outperforming  state-of-the-art methods on the MIT Indoor67 dataset~\cite{quattoni2009recognizing}.

\end{itemize}
Details on the architecture, training set and layer sizes of the CNNs are summarized in Table~\ref{fig:cnns}.

\begin{table}[ht]
\caption{Details on architecture, training set and layer size of the CNNs.}
\label{fig:cnns} 
\centering
\ra{1.2}
{\footnotesize \singlespacing
\begin{adjustbox}{max width=\textwidth,center}
\begin{tabular}{@{}lrrrcrrrcrrrr@{}}
\toprule
& \multicolumn{3}{c}{\sc Architecture} && \multicolumn{3}{c}{\sc Training} && \multicolumn{4}{c}{\sc Layer Size}\\
\cmidrule{2-4} \cmidrule{6-8} \cmidrule{10-13}
& parameters & depth (conv+fc) & input size && training set & classes & data size && pool5 & fc6 & fc7 & fc8\\
\midrule
{\it OxfordNet} & 138M & 13+3 & $224\times224\times3$ && ImageNet & 1000 & 1.2M && $7\times7\times512$ & 4096 & 4096 & 1000\\
{\it AlexNet} & 60M & 5+3 & $227\times227\times3$ && ImageNet & 1000 & 1.2M && $6\times6\times256$ & 4096 & 4096 & 1000\\
{\it PlacesNet} & 60M & 5+3 & $227\times227\times3$ && Places-205 & 205 & 2.4M && $6\times6\times256$ & 4096 & 4096 & 205\\
{\it HybridNet} & 60M & 5+3 & $227\times227\times3$ && Both & 1183 & 3.6M && $6\times6\times256$ & 4096 & 4096 & 1183\\
\bottomrule
\end{tabular}
\end{adjustbox}
}
\end{table}

These state-of-the-art models are chosen as they allow us to run interesting control experiments, where the CNN architecture or training data are varied.
{\it PlacesNet} and {\it HybridNet} share the same architecture as {\it AlexNet}~\cite{Zhou2014}, while being trained on different data.
{\it OxfordNet} and {\it AlexNet} are trained on the same data, but have different architectures: compared to {\it AlexNet}, {\it OxfordNet} is deeper, has twice as many layers, twice the number of parameters, and achieved better image classification performance in the ImageNet 2014 contest~\cite{Simonyan2014}.

The 4 models are trained differently, using the ImageNet~\cite{Deng2009} and Places-205~\cite{Zhou2014} datasets. 
With categories like ``Amphitheater'', ``Jail cell'' or ``Roof garden'', Places-205 is a scene-centric dataset, while ImageNet, featuring categories such as ``Vending machine'', ``Barn spider'' or ``Chocolate syrup'', is more object-centric.
Places-205 is twice as large as ImageNet, but has 5 times fewer classes.
{\it OxfordNet} and {\it AlexNet} are trained on ImageNet. 
{\it HybridNet} is trained on a combination of ImageNet and Places-205 data: the resulting dataset being 3 times larger than ImageNet alone, and having a larger variety of classes.

Given an input image, we first resize it to a canonical resolution, compute the feed-forward neural network activations, and extract the last four layers for each CNN model. 
We refer by $pool5$, $fc6$, $fc7$ and $fc8$ outputs of the last 4 layers of each network (as denoted in Caffe).
$pool5$ is the output of the last convolutional layer after pooling, and $fc6$, $fc7$, $fc8$ are outputs of the fully connected layers.
$pool5$ still contains spatial information from the input image.
The size of the last layer $fc8$ is equal to the number of classes. 
All descriptors are extracted after applying the rectified linear transform, and $L_2$ normalized: the features are directly output from the Caffe implementation of the CNN models.

\section{Experimental Results}
\label{sec:experiments}

\begin{figure}[ht]
	\centering{
		\includegraphics[width=0.8\textwidth]{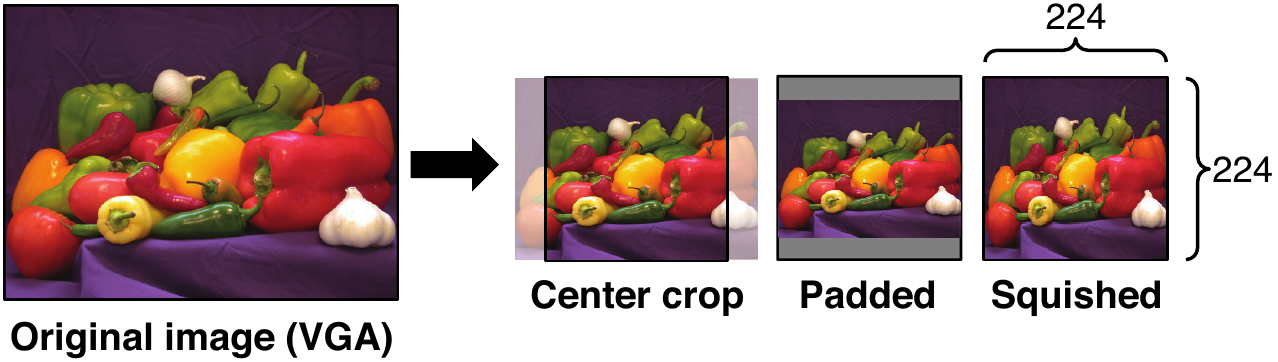}
		\caption{Different single-crop strategies used for input into CNN pipelines.
	}
	\label{fig:cnn_cropping_strategy}
	}
\end{figure}

\begin{figure}[ht]
	\centering{
		\includegraphics[width=0.49\textwidth]{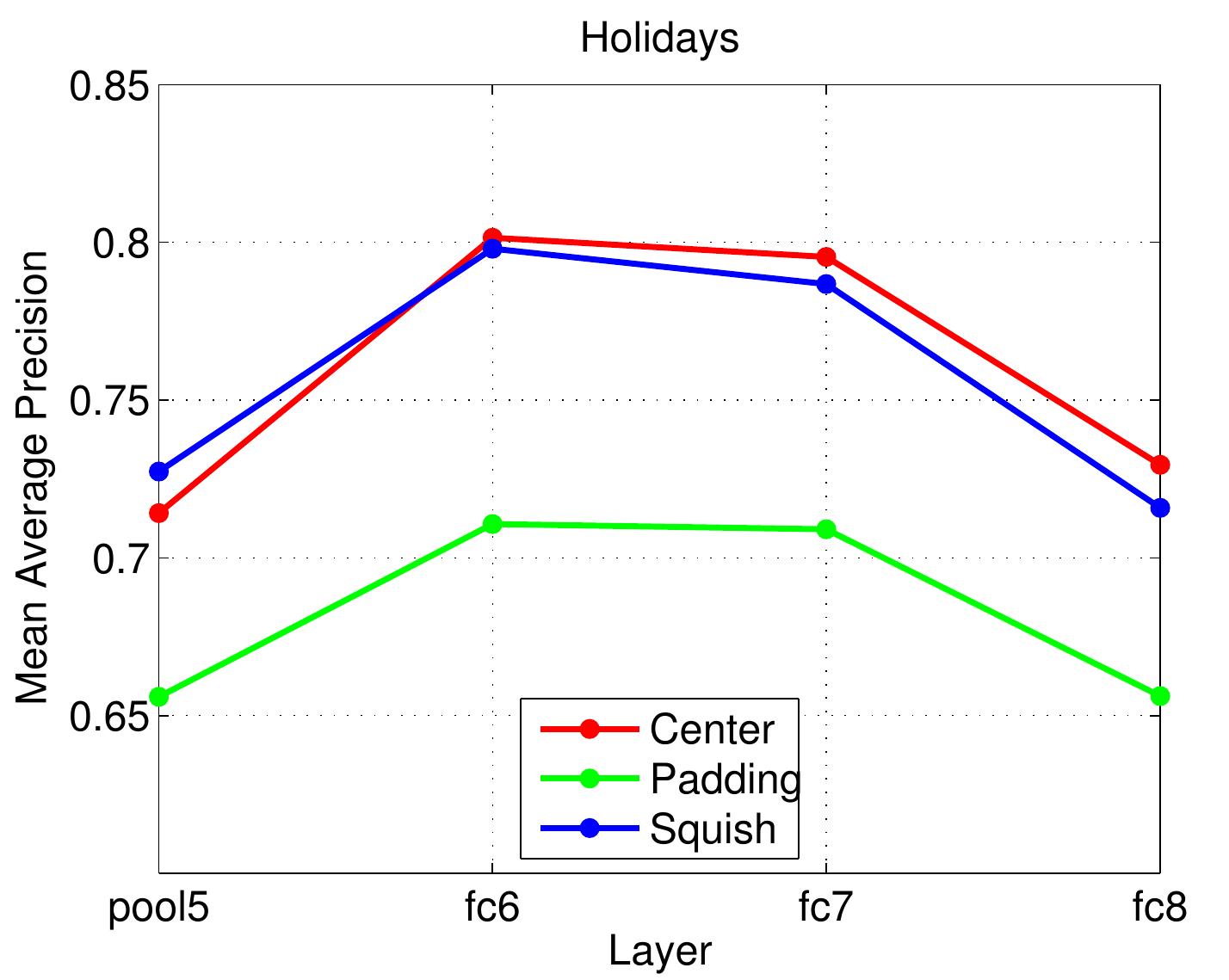}
		\caption{MAP for different layers of {\it OxfordNet}, for different single-crop strategies on the {\it Holidays} data set. We observe that {\it Center} crop and {\it Squish} perform comparably.
	}
	\label{fig:layer_strategy}
	}
\end{figure}

\subsection{Best practices for CNN descriptors}

\textbf{What single crop strategy is the best?} 

CNN pipelines take input images at a fixed resolution (see Table~\ref{fig:cnns}).
We wish to determine which single-crop strategy works best in the context of instance retrieval where images may vary in size and aspect ratio.
We consider the following 3 different cropping strategies illustrated in Figure~\ref{fig:cnn_cropping_strategy}.
Numerical values are given to fit {\it OxfordNet}.
\begin{itemize}
\item {\it Center}: the largest 224 $\times$ 224 center crop, after rescaling the image to 224 pixels for the smaller dimension, while maintaining aspect ratio.
\item {\it Padding}: the original image is resized to 224 pixels for the larger dimension, maintaining aspect ratio, and any unfilled pixels are padded with a constant value equal to the training set mean.
\item {\it Squish}: the original image is resized to 224 $\times$ 224. The original aspect ratio is ignored potentially resulting in distortions.
\end{itemize}
In Figure~\ref{fig:layer_strategy}, we plot MAP for different layers of {\it OxfordNet}, for the {\it Holidays} data set.
We note that {\it Center} and {\it Squish} perform comparably, outperforming {\it Padding}. 
The trend is consistent across the different network layers.
We observe similar results for other data sets and CNN models.
Most data sets in this study have a center bias for the object of interest, explaining the best performances of the {\it Center} cropping strategy.

A small improvement in performance for large-scale image classification is obtained by averaging output class probabilities computed over several cropped regions within an image, often extracted at different positions and scales~\cite{VeryDeepNeuralNets}.
Such a performance improvement can also be achieved for instance-retrieval by pooling CNN results over several cropped regions, but such a strategy could be applied to other global descriptor pipelines too.
For the remaining experiments in this paper, we consider a single {\it Center} crop for processing all database and query images.

\textbf{Which CNN layer performs the best?}

\begin{figure*}[ht]
\centering{
	\begin{tabular}{@{}c@{} @{}c@{}}
		\includegraphics[width=0.49\textwidth]{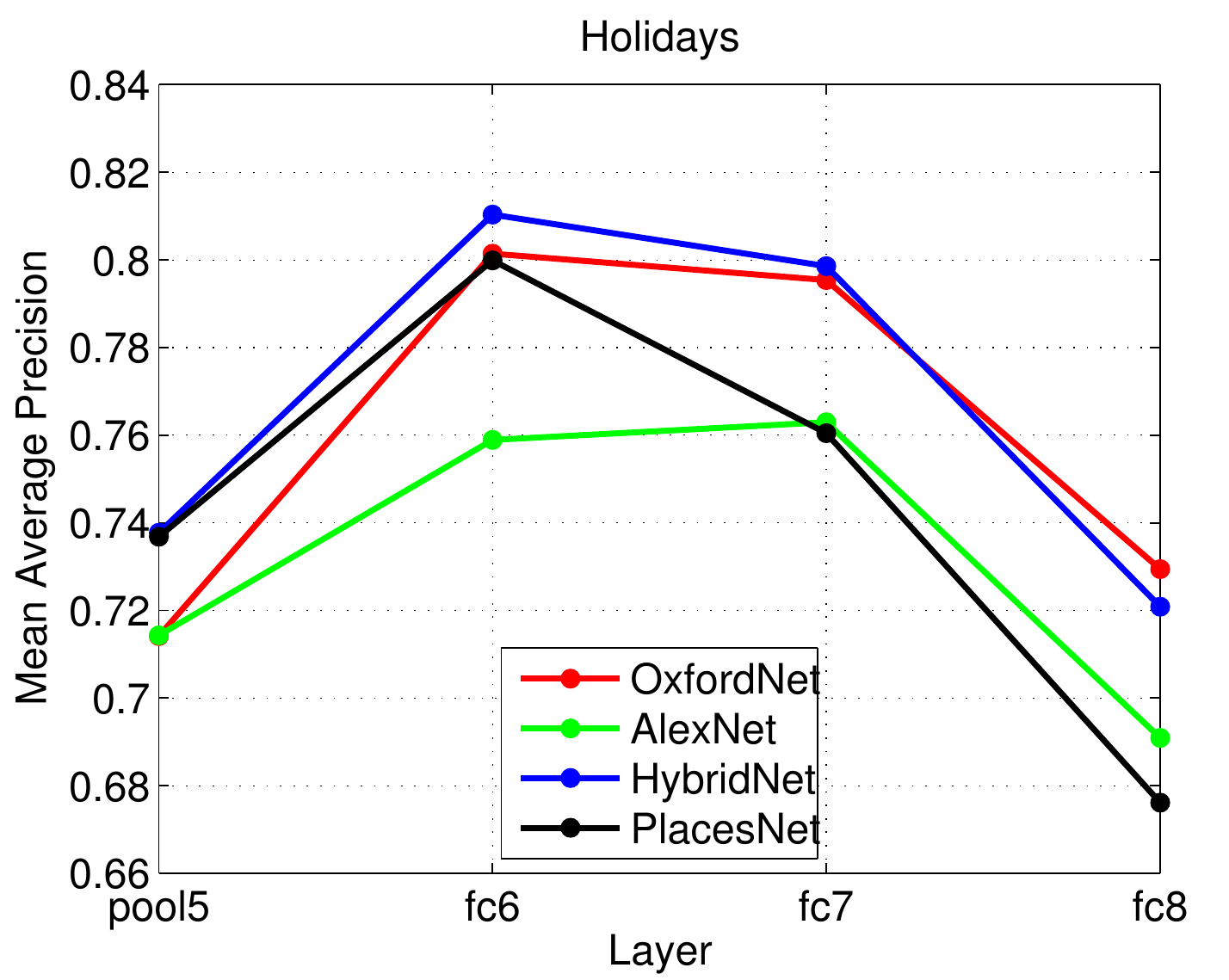} &
		\includegraphics[width=0.49\textwidth]{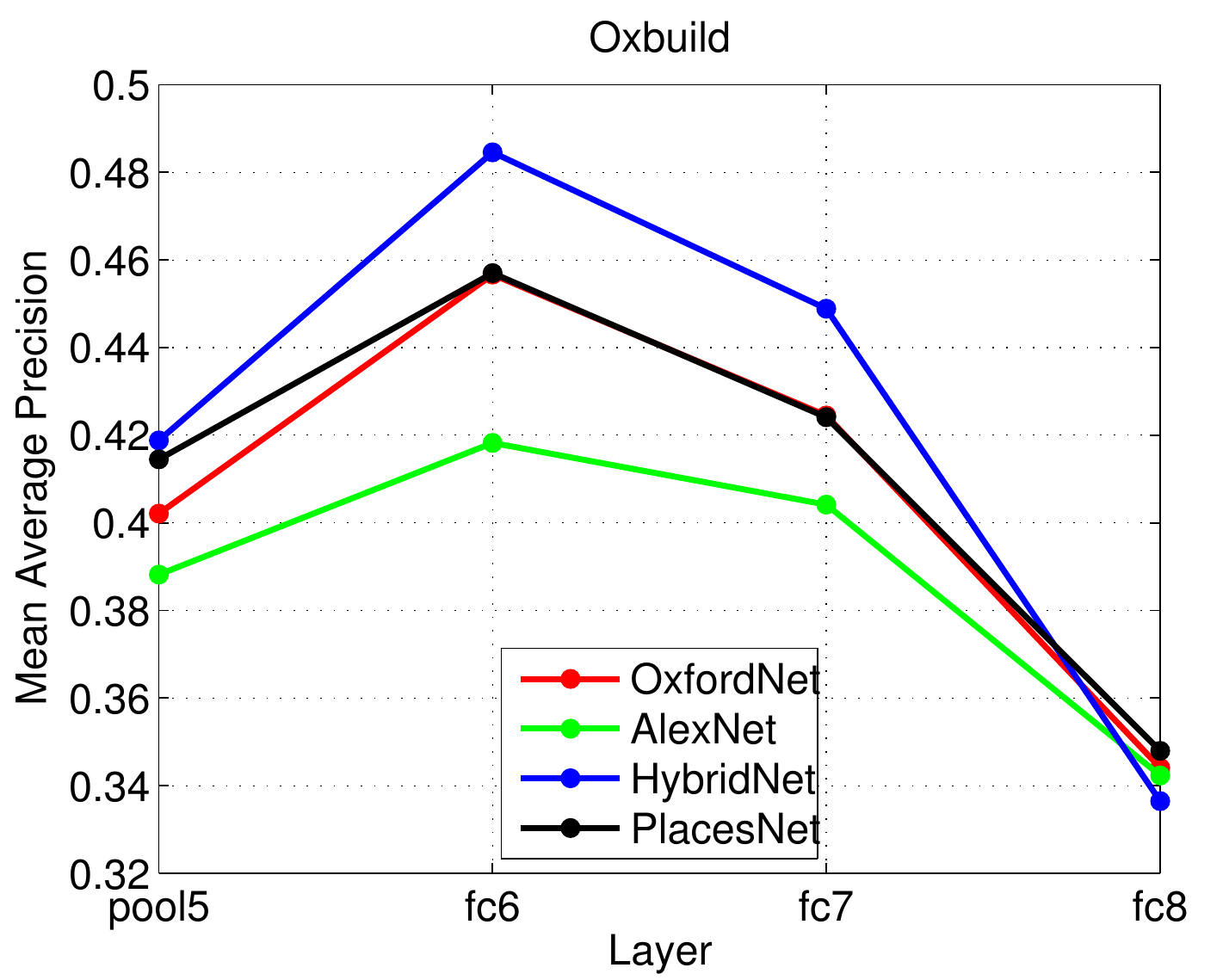} \\
		\includegraphics[width=0.49\textwidth]{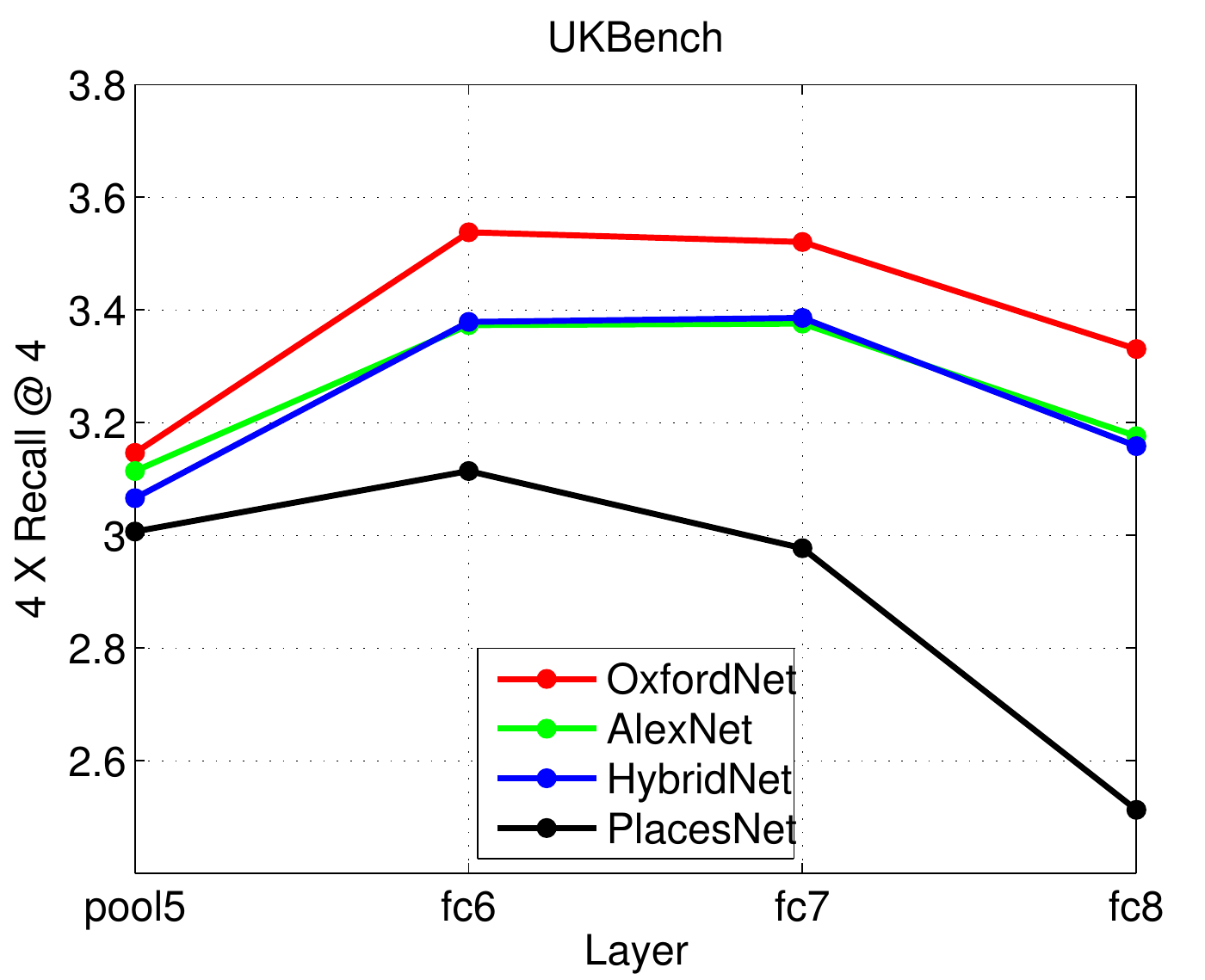} &
		\includegraphics[width=0.49\textwidth]{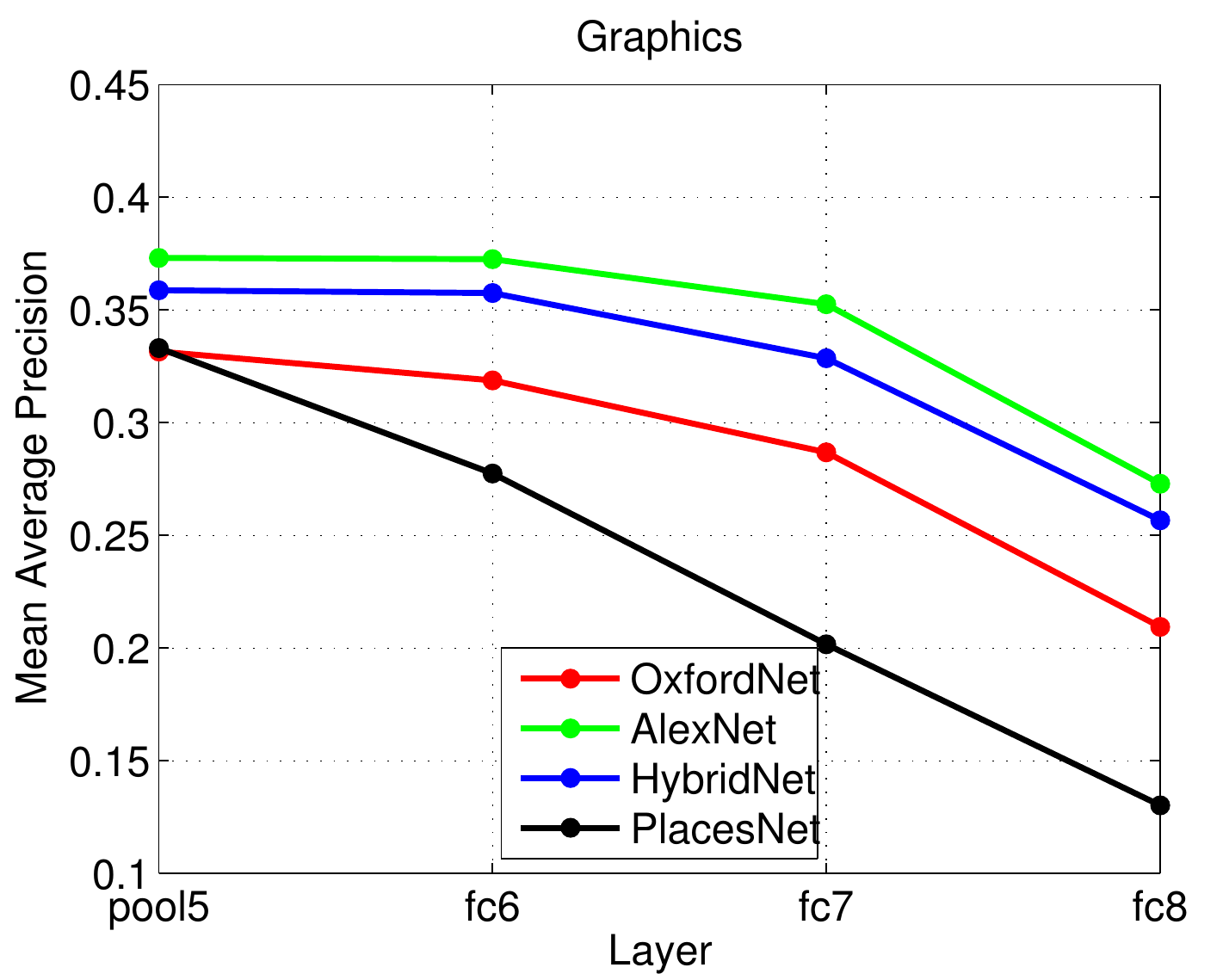} \\
	\end{tabular}
	\caption{MAP for the last 4 layers of state-of-the-art publicly available CNN. {\it OxfordNet} and {\it AlexNet} are trained on the same data, while {\it PlacesNet}, {\it HybridNet} and {\it AlexNet} have the same network architecture but are trained on different data. We note that performance improves by using deeper networks, and by training on domain specific data, but only if training and testing data have similar characteristics.
		}
\label{fig:best_layer}
}	
\end{figure*}

In Figure~\ref{fig:best_layer}, we plot MAP for the last 4 layers of {\it OxfordNet}, {\it AlexNet}, {\it PlacesNet} and {\it HybridNet} for different data sets.
We note that for each network,  intermediate layers perform best for instance retrieval.
Such a sweet spot is intuitive as the final layer represents higher level semantic concepts, while intermediate convolutional and fully connected layers provide rich representations of low level image information.
We note that layer $fc6$ performs the best for all CNN, for all data sets except {\it Graphics}.
For {\it Graphics}, performance drops with increase in depth, as all four CNN models are learnt on natural image statistics, while the {\it Graphics} data set is biased towards data like CD covers, DVD covers, business cards, and dense text in newspaper articles.

\textbf{How much improvement can we obtain by deeper CNN architectures?}

We compare {\it OxfordNet} and {\it AlexNet} results in Figure~\ref{fig:best_layer}.  
{\it OxfordNet} and {\it AlexNet} are both trained on the same 1.2 million images from the ImageNet data set, but vary in the number of layers: 16 and 8 layers respectively.
We note that {\it OxfordNet} outperforms {\it AlexNet} on all data sets, except {\it Graphics}.
On {\it Graphics}, the performance of {\it OxfordNet} is worse, strongly suggesting that performance improves with more layers as long as the training data is representative of the test data set in consideration.

\textbf{How much improvement can we obtain by training CNN models using domain specific data?}

For this experiment, we compare {\it AlexNet} with {\it PlacesNet} in Figures~\ref{fig:best_layer}.
{\it AlexNet}, and {\it PlacesNet}  use the same 8-layer CNN architecture, but are trained on different data.
We observe that {\it PlacesNet} outperforms {\it AlexNet} on {\it Holidays} and {\it Oxbuild} data sets. 
This shows that using training data more representative of the test data can improve performance significantly, as {\it Holidays} and {\it Oxbuild} are scene-centric.
On the object-centric {\it UKBench} and {\it Graphics} data sets, {\it PlacesNet} performs worse than {\it AlexNet} due to the mismatch between training and test data.

Further, in Table~\ref{tab:cnn_fv_results}, we compare our results to the CNN retrieval results presented in~\cite{Yandex}.
In~\cite{Yandex}, Babenko et al. fine-tune a pre-trained {\it AlexNet} model based on ImageNet training data with domain specific images, e.g., landmarks and objects.
As shown in Table~\ref{tab:cnn_fv_results}, the authors are able to improve retrieval performance over the {\it AlexNet} baseline model, on {\it Holidays} and {\it UKBench} by fine-tuning with landmark and object data respectively.
However, the resulting trade-off is a loss in performance on {\it Holidays} when fine-tuning with object data and vice-versa.

We compare {\it OxfordNet} results with those of the fine-tuned models of~\cite{Yandex}.
We note that the deeper architecture of {\it OxfordNet} trained on just ImageNet data results in comparable or higher performance than the fine-tuned models on both {\it Holidays} and {\it UKBench}, suggesting that there is more gain to be had with deeper networks rather than fine-tuning a shallower network with domain-specific data.

Together with the previous sets of experiments, there is strong combined evidence that deeper layers/models have the potential of achieving higher discriminativeness on domain specific data at the expense of less generalisability on non-specific data.

\textbf{How much improvement can we obtain by training CNN on larger and more diverse data?}

For this experiment, we compare {\it AlexNet} with {\it HybridNet} in Figure~\ref{fig:best_layer}.
{\it AlexNet}, and {\it HybridNet}  use the same 8-layer CNN architecture, but the latter is trained on a combination of ImageNet and Places-205, resulting in a larger training data set with more diverse classes.
We note that {\it HybridNet} performs comparably or better than {\it AlexNet} on all data sets except {\it Graphics}, suggesting that increasing the amount and diversity of training data is equally important as increasing depth in the CNN architecture.

\begin{table}[ht]
\caption{State-of-the-art CNN and FV results for instance retrieval. 4 $\times$ Recall $@ $ 4 for {\it UKBench}, and MAP for other data sets.}
\label{tab:cnn_fv_results}
\centering
{\footnotesize \singlespacing
\begin{adjustbox}{max width=\textwidth,center}
\begin{tabular}{@{}lrrrrr@{}}
\toprule
Descriptor & Dim & Holidays & UKBench	& Oxbuild 	& Graphics\\
\midrule
{\it OxfordNet} & 4096 & 0.80 & 3.54	& 0.46 & 0.33\\
{\it AlexNet} & 4096 & 0.76 & 3.38 &	0.42 	& 0.37\\
{\it HybridNet} & 4096 & 0.81 & 3.39	& 0.48 & 0.36\\	
{\it PlacesNet}	& 4096 & 0.80 & 3.11 & 0.46 & 0.33\\
CNN (Fine-tuned &&&&&\\
on Landmarks)~\cite{Yandex}  &	4096	& 0.793 & 3.29	& 0.545 &\\
CNN (Fine-tuned &&&&&\\
on Objects)~\cite{Yandex} & 4096 & 0.754 & 3.56 & 0.393 & \\
\midrule
FVDoG								&	32768	&	0.63	&	2.8		&	0.42	& 0.66 \\
FVDS								&	32768	&	0.73	&	2.38	&	0.51	& 0.20 \\
FVDM								&	32768	&	0.75	&	2.45	&	0.55	& 0.32 \\
\bottomrule
\end{tabular}
\end{adjustbox}
}
\end{table}

\subsection{Best practices for FV interest points}

CNN features are obtained by dense sampling over the image.
In Table~\ref{tab:cnn_fv_results}, we study if such an approach is also effective for FVs.
We compare the performance of FVDoG, FVDS and FVDM as described in Section~\ref{sec:eval_framework}.

We note that dense sampling (FVDS and FVDM) improves performance over FVDoG on {\it Holidays} and {\it Oxbuild} data sets, while hurting performance on {\it Graphics} and {\it UKBench}.
Note the large drop in performance of dense sampling on the {\it Graphics} data set.
This is intuitive as queries in the {\it Graphics} data set contain query objects at different scales and rotations. 
For {\it Holidays} and {\it Oxbuild} data sets, even FVDS improves performance over FVDoG, suggesting that most query and database image pairs occur at roughly the same scale.

Dense sampling is effective for data sets like {\it Holidays} which consist primarily of outdoor scenes, and are mainly composed of highly textured patches.
The improvement in performance of dense sampling approaches can also be attributed to the discriminativeness-invariance tradeoff.
Where retrieval does not require scale and rotation invariance, and data are highly textured over the entire image, performance can be improved by dense sampling. 

Sampling at multiple scales also seems to consistently improve results over single scale sampling for dense descriptors.

\subsection{Comparisons to state-of-the-art}

\textbf{Does combining FV and CNN improve performance?} 

In Figure~\ref{fig:combine_fv_cnn}, we present retrieval results obtained by combining FVDoG, FVDS, and FVDM individually with {\it OxfordNet} {\it fc6} features.
We employ a simple early fusion approach where the FV and CNN features are concatenated after weighting by $\alpha$ and $(1-\alpha)$ respectively.
$\alpha=0$ corresponds to using just FVDoG, FVDS or FVDM features individually, while $\alpha=1$ corresponds to just the {\it OxfordNet} feature.
This early fusion scheme is also equivalent to weighting the squared $L_2$ distance measure for matching by $\alpha$ and $1-\alpha$ for FV and CNN features respectively.

All four data sets show an improvement in peak performance by combining FV and CNN features.
The maximum performance is achieved for $\alpha=0.4$ for the {\it Holidays}, {\it UKBench} and {\it Oxbuild} data sets, and $\alpha=0.3$ for the {\it Graphics} data set, using different FV.
There is a significant improvement in performance by combining FV and CNN features on all data sets except {\it Graphics}.
The results suggest that a simple hyperparameter can be used to combine FV and CNN across data sets with similar characteristics.
Also, $\alpha=0.4$ suggests that FV contribute significantly to achieving high retrieval performance (in contrast, an $\alpha$ parameter close to 1 would suggest that most of the contribution is from the CNN feature).

Note that our goal here is to show that performance can be improved significantly by combining FV and CNN features, and not necessarily to achieve highest performance on these retrieval benchmarks.
Peak performance presented in Figure~\ref{fig:combine_fv_cnn} can be improved by (a) database-side rotation and scale pooling which helps significantly (see Sections~\ref{sec:rotation_experiments} and \ref{sec:scale-experiments}) (b) better CNN models than {\it OxfordNet} on individual data sets (see Table~\ref{tab:cnn_fv_results}) (c) better FV based on Hessian Affine interest points~\cite{HessianSIFTCode} instead of DoG points used in this paper~\cite{VidaldiSIFT2} (d) better FV with more sophisticated aggregation techniques~\cite{TEDA}, and (e) combining all FV and CNN descriptors together.
(f) using more sophisticated fusion and ranking techniques for combining results, like the one proposed in the recent paper~\cite{fusionretrieval}.

\begin{figure}[ht]
\centering{
	\begin{tabular}{@{}c@{} @{}c@{}}

		\includegraphics[width=0.49\textwidth]{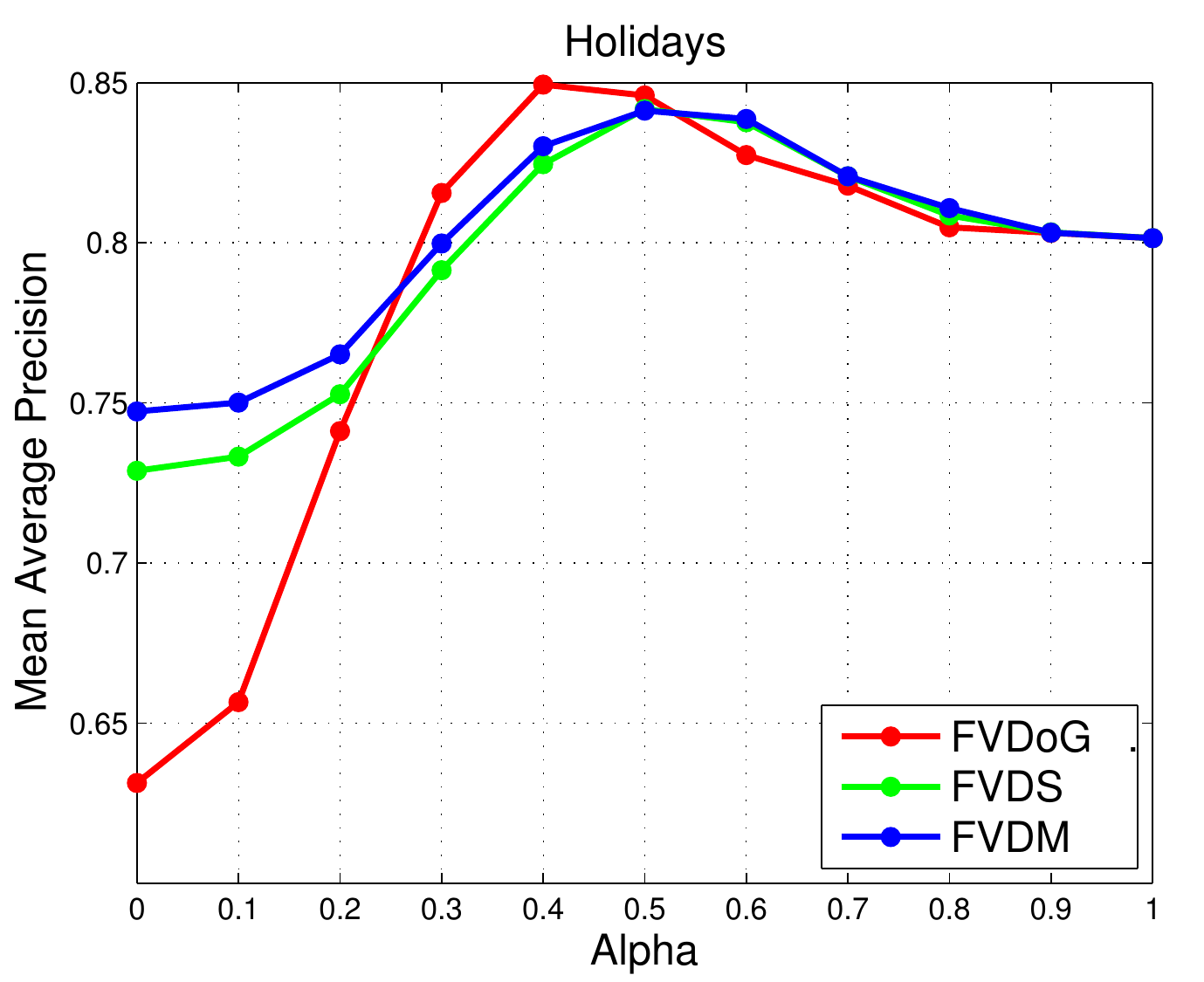} &
		\includegraphics[width=0.49\textwidth]{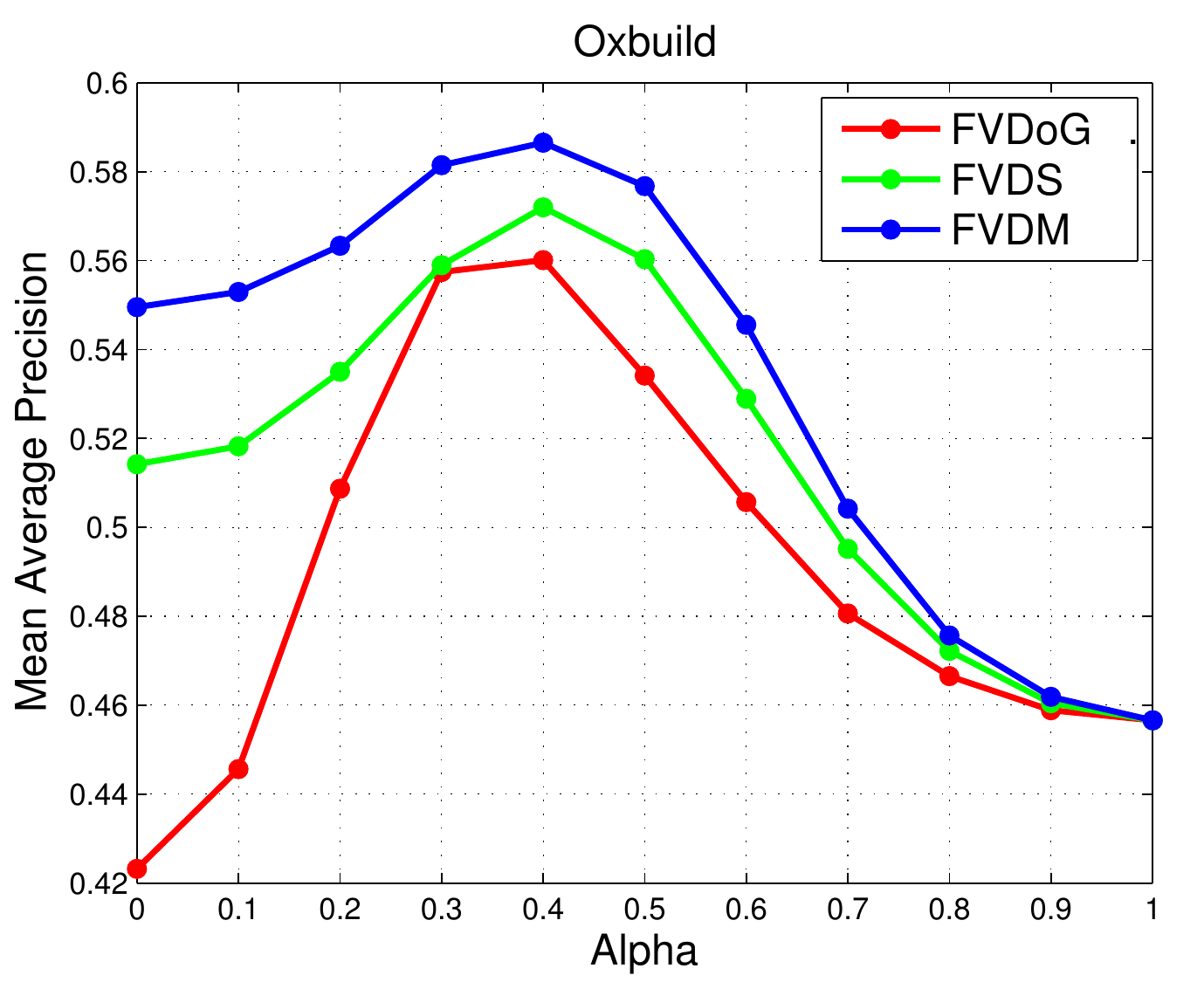} \\
		\includegraphics[width=0.49\textwidth]{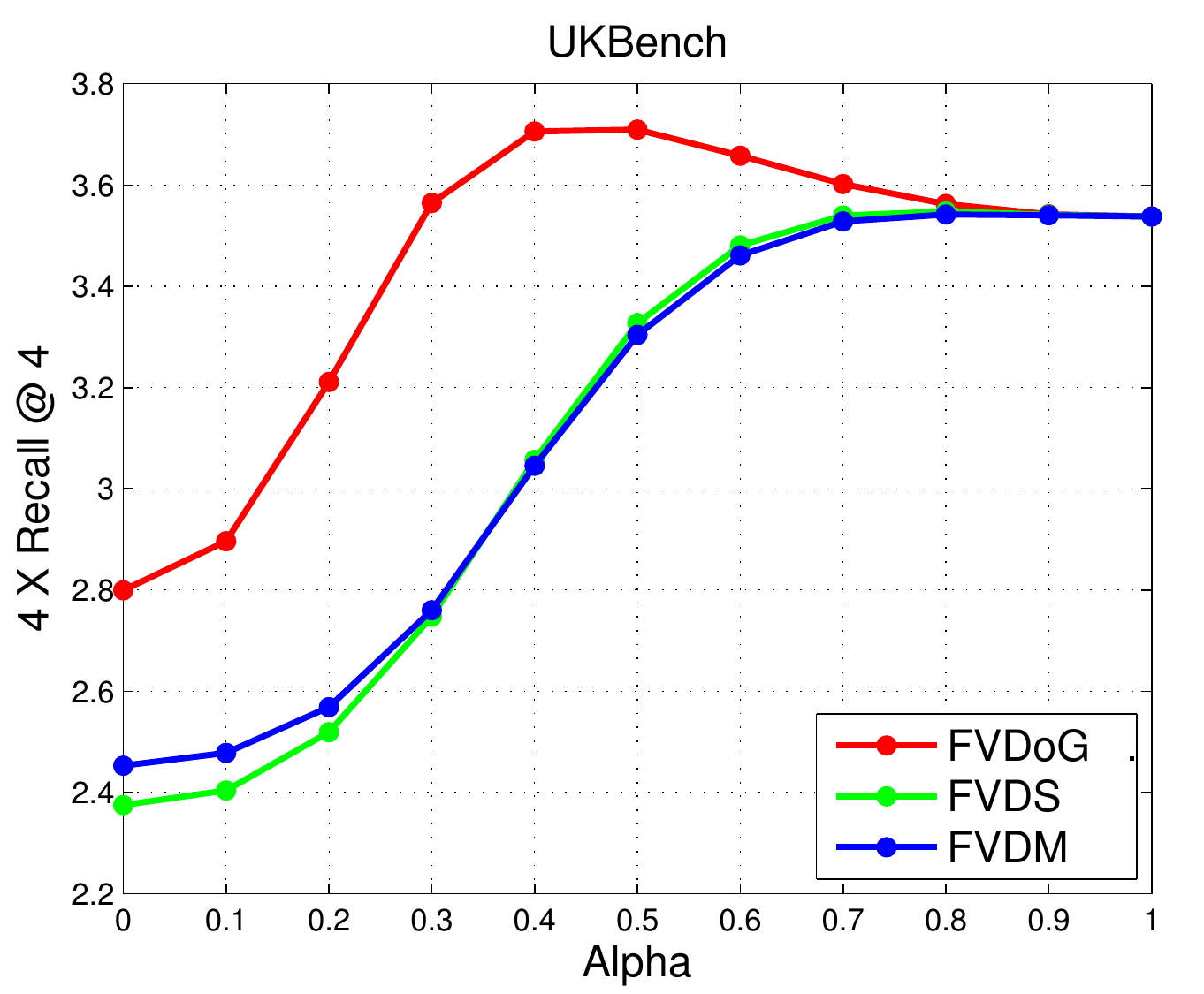} &
		\includegraphics[width=0.49\textwidth]{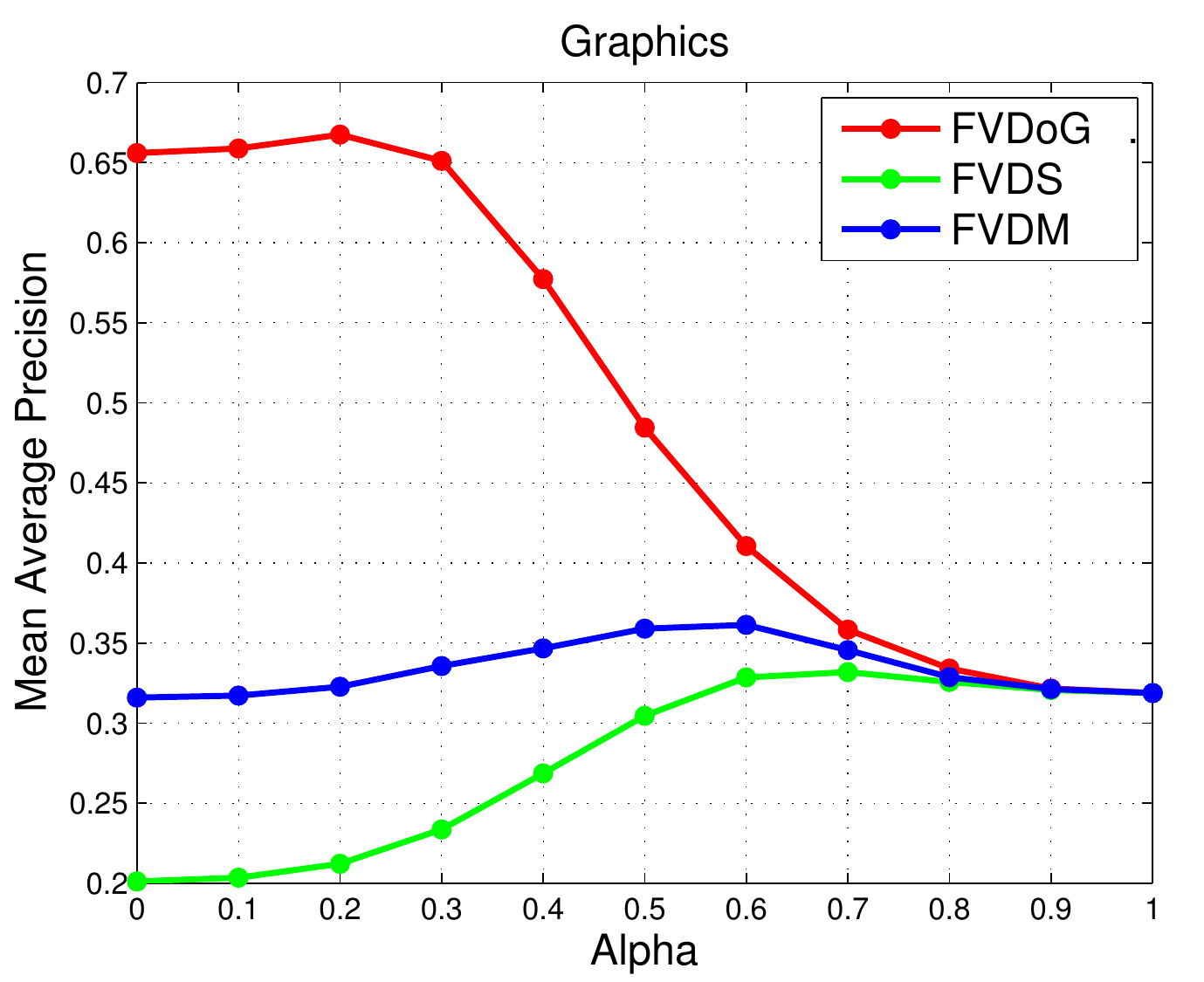} \\

	\end{tabular}
	\caption{Combining different FVs with {\it OxfordNet} {\it fc6} with early fusion. FV and {\it OxfordNet} features are concatenated with weights $\alpha$ and $1-\alpha$ respectively.  $\alpha=0$ refers to just using FVDoG, FVDS or FVDM above, while $\alpha=1$ refers to just the {\it OxfordNet} {\it fc6} feature. We observe that retrieval performance improves on all data sets by combining FV and CNN.
		}
\label{fig:combine_fv_cnn}
}	
\end{figure}

\begin{table}[ht]
\caption{State-of-the-art results. MAP for {\it Oxbuild} and {\it Holidays}, and $4 \times$ Recall $@ 4$ for {\it UKBench}}
\label{tab:final_results}
\centering
{\footnotesize \singlespacing
\begin{adjustbox}{max width=\textwidth,center}
\begin{tabular}{@{}lrrrr@{}}
\toprule
Descriptor 							&	Dim 	& Holidays		& UKBench			& Oxbuild \\
\midrule
Bag-of-words 1M~\cite{Nister06}		&	1M		&			&	3.19		&	\\

VLAD baseline~\cite{Jegou_CVPR_10}	& 8192		&	0.526	&	3.17		&	\\ 

Fisher baseline~\cite{Jegou_CVPR_10}& 8192		&	0.495	&	3.09		&	\\ 

Fisher baseline (ours)				&	32768	&	0.63	&	2.8			&	0.42 \\

Fisher ADC (320 bytes)~\cite{PQFisher}& 2048	&	0.634	&	3.47		&	\\ 

Fisher+color~\cite{FisherColor}		& 4096		&	0.774	&	3.19		&	\\

VLAD++~\cite{AllAboutVLAD}			& 32768		&	0.646	&				& 0.555 \\

Sparse-coded features~\cite{SCF}	& 11024		&	0.767	&	{\textbf{3.76}}		& \\

Triangulation Embed~\cite{TEDA}	& 8064		&	0.77	& 				& {\textbf{0.676}} \\

Triangulation Embed~\cite{TEDA}	& 1920		&			&  	3.53		&  \\

Best CNN results					& 4096		&	0.81	&	3.54		& 0.48	\\
from this paper						& 			&			&				& 	\\
across all CNN						& 			&			&				& 	\\

Fusion of {\it OxfordNet} and		& 32768+ &	{\textbf{0.85}}	&	3.71		& 0.59	\\
Baseline FV							& 	4096		 &			&				& 		\\
\bottomrule
\end{tabular}
\end{adjustbox}
}
\end{table}

\textbf{Comparisons to state-of-the-art.} 

In Table~\ref{tab:final_results}, we compare state-of-the-art results reported on {\it Holidays}, {\it UKBench} and {\it Oxbuild}.
We include a wide range of approaches starting from Bag-of-words~\cite{Nister06} to latest FV aggregation methods~\cite{TEDA}.
We include the best CNN and fusion results reported in this paper.
We note that the best CNN results (based on pre-trained models considered in this work) achieve higher performance than state-of-the-art FV approaches~\cite{TEDA} on {\it Holidays} and {\it UKBench} data sets.
There is a gap in performance between CNN results reported in this work and state-of-the-art FV for {\it Oxbuild}: however, {\it Oxbuild} is a much smaller data set with only 55 queries. 
Finally, we note that the simple fusion technique in Figure~\ref{fig:combine_fv_cnn} results in highest or one of the highest performance numbers reported on each data set.
Peak performance numbers for the fusion approach can be improved using approaches (a)-(f) described above.

\subsection{Invariance to Rotation}
\label{sec:rotation_experiments}

\begin{figure}[ht]
	\centering{
		\includegraphics[width=0.7\textwidth]{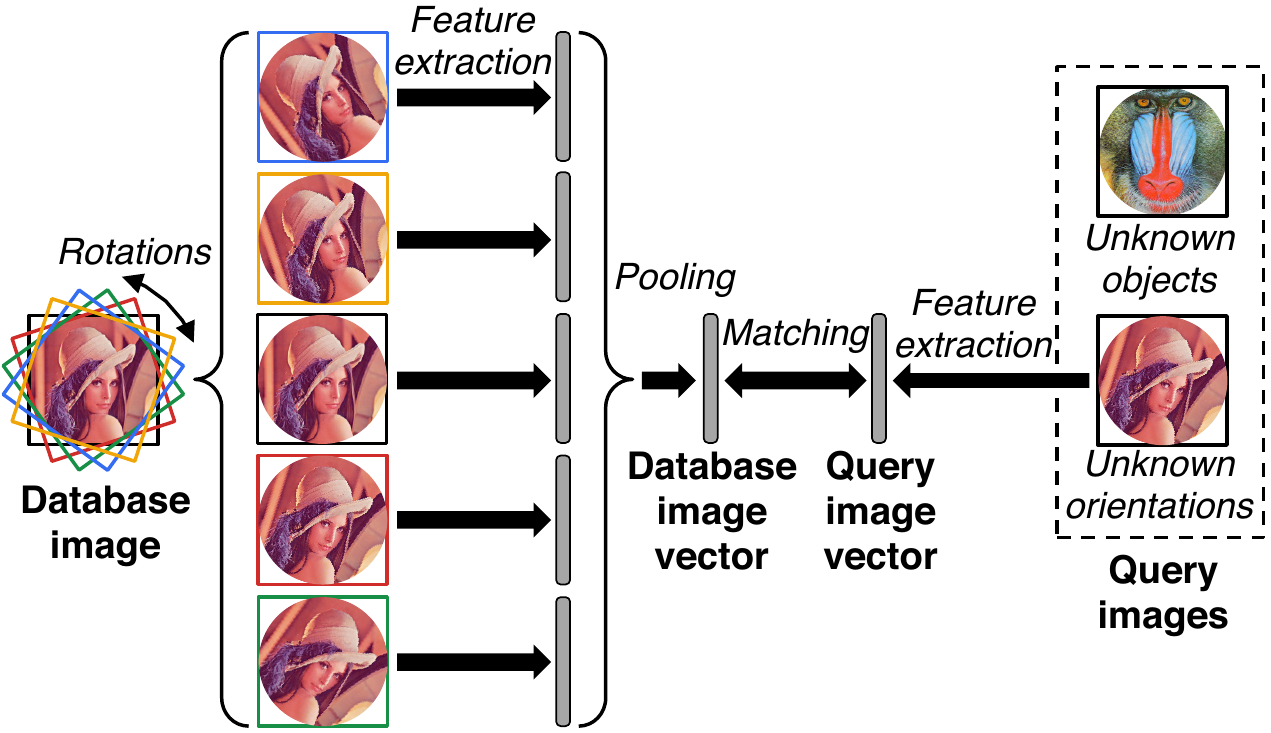}
		\caption{We extract features from database images at different rotations and pool them to obtain a single representation. We rotate queries and evaluate retrieval performance for different pooling parameters and strategies. All database and query images are cropped circularly at the center to avoid edge artifacts for this experiment, and padded with a default mean RGB value (ImageNet mean).
	}
	\label{fig:cnn_rot_illus}
	}
\end{figure}

\textbf{How invariant are CNN features to rotation?}

\begin{figure}[ht]
	\centering{
		\includegraphics[width=0.49\textwidth]{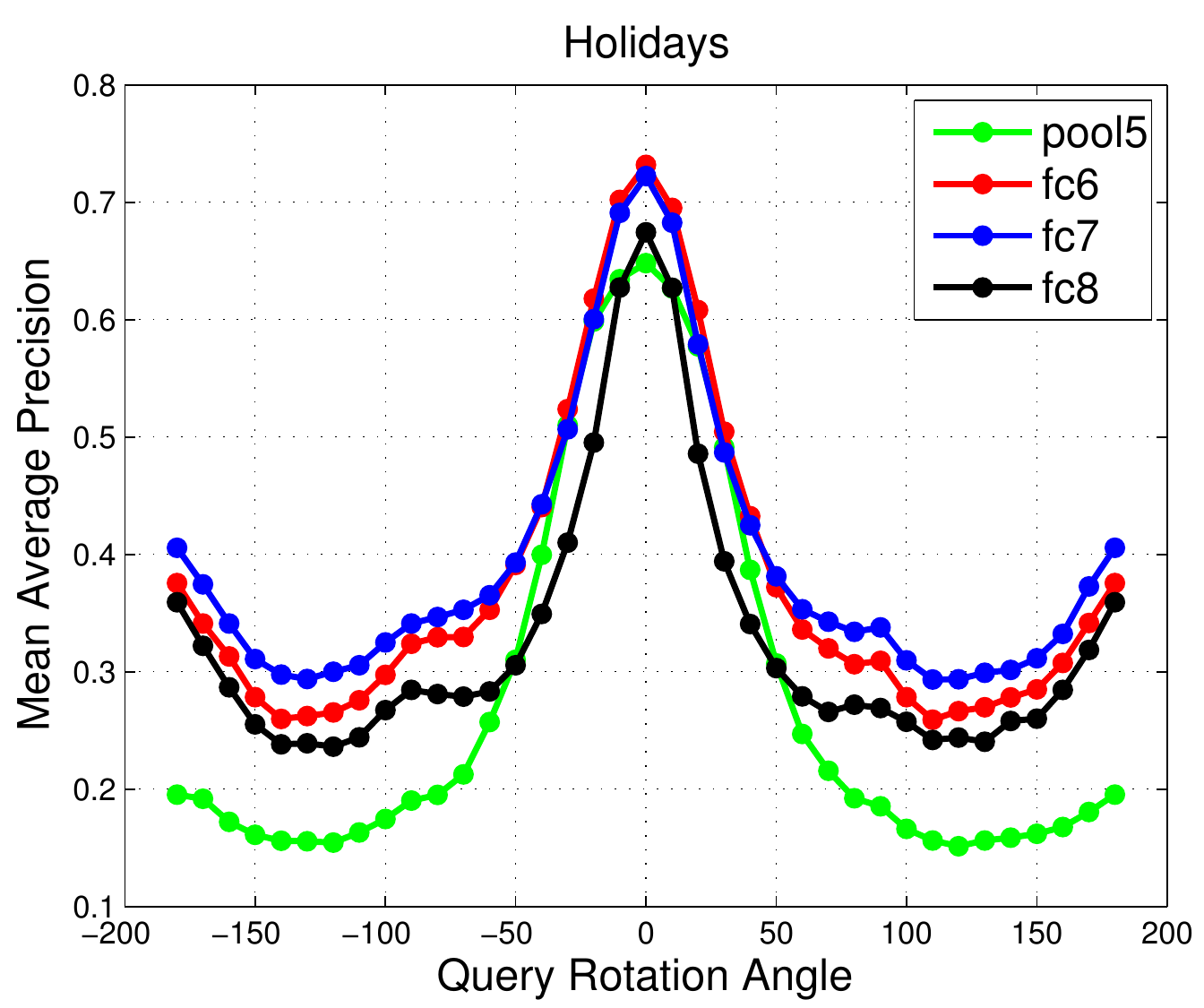}
		\caption{MAP as query images are rotated for different layers of {\it OxfordNet}. We note that CNN features have very limited rotation invariance, with performance dropping steeply for all layers of the network beyond 10$\degree$.
	}
	\label{fig:cnn_rot_invariance}
	}
\end{figure}

\begin{figure}[ht]
\centering{
	\begin{tabular}{@{}c@{}@{}c@{}}
		\includegraphics[width=0.49\textwidth]{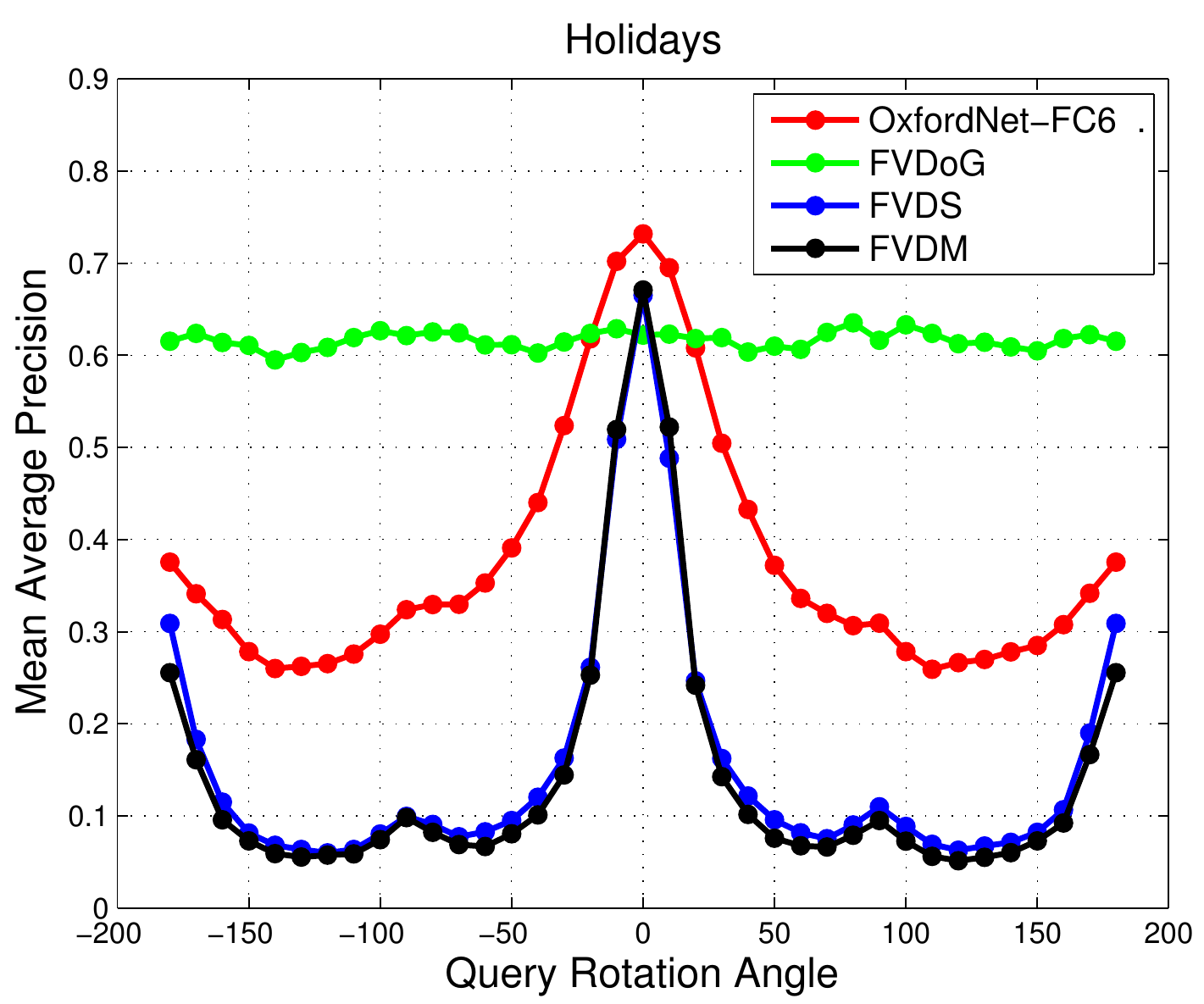} &
		\includegraphics[width=0.49\textwidth]{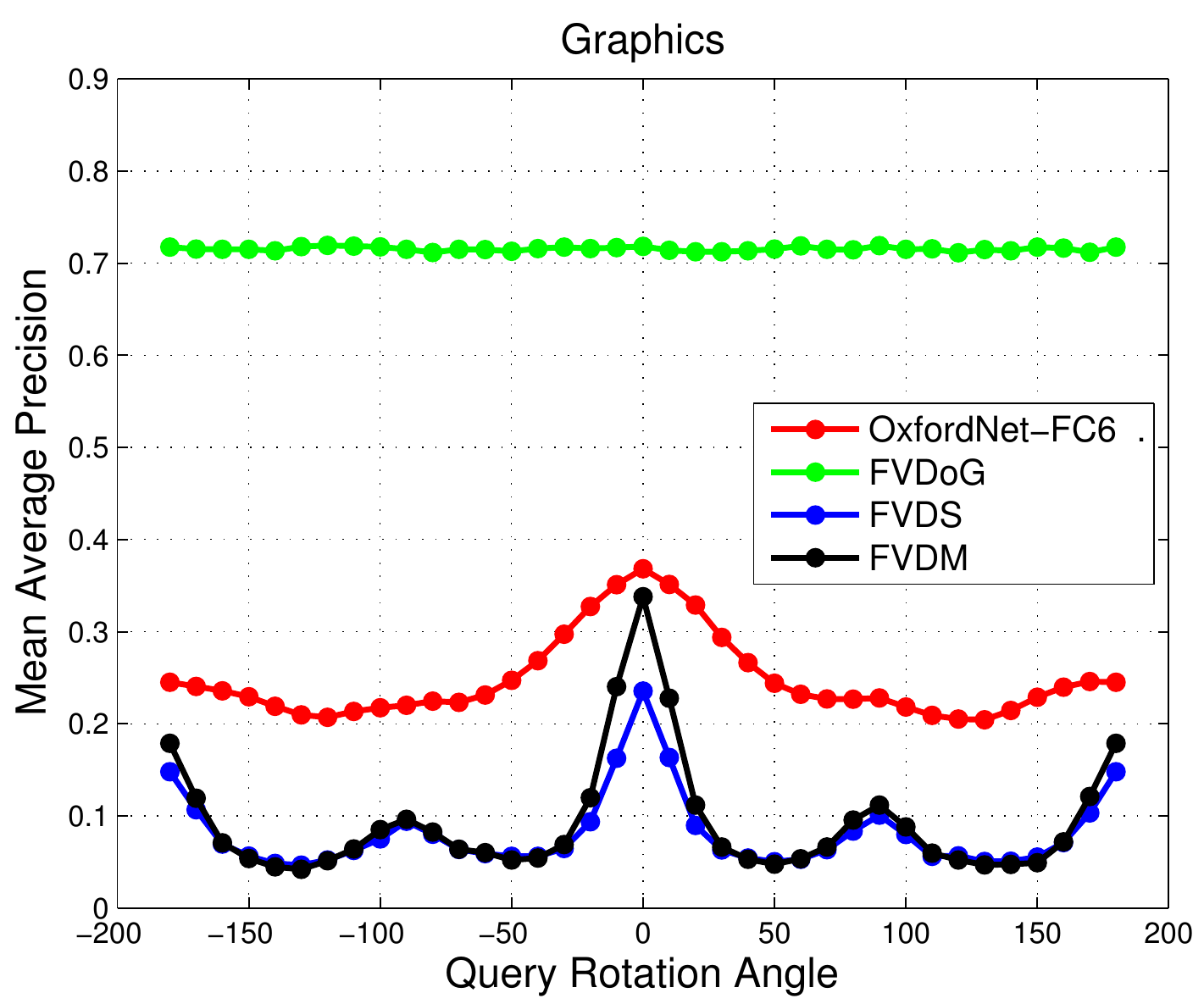}
	\end{tabular}
	\caption{Comparison of {\it OxfordNet} {\it fc6} and FV for rotated queries on the {\it Holidays} and {\it Graphics} data sets. The FVDoG is robust to rotation, while {\it OxfordNet}, FVDS, FVDM suffer a sharp drop in performance.
		}
\label{fig:cnn_vs_fisher_rotation}
}	
\end{figure}

CNN features, unlike FVDoG, have limited levels of rotation invariance.
The invariance arises from the max-pooling steps in the CNN pipeline, and rotated versions of objects present in the training data.
In Figure~\ref{fig:cnn_rot_invariance}, we rotate each query at different angles and measure MAP for the {\it Holidays} data set for different layers of {\it OxfordNet}.
For these control experiments, query images are cropped circularly in the center (to avoid edge artifacts) and rotated in steps of 10$\degree$. 
The same experimental set up is also employed in evaluation of rotation invariant features in~\cite{RIFF}.
We note that CNN features have very limited rotation invariance with performance dropping steeply beyond 10$\degree$.
Furthermore, the different layers of the network exhibit similar characteristics suggesting that rotation invariance does not increase with depth in the CNN.

\textbf{Are FVs or CNNs more rotation invariant?}

For  the sake of evaluation, we choose one scene-centric and one object-centric data set that are most different: {\it Holidays} and {\it Graphics}.
In Figure~\ref{fig:cnn_vs_fisher_rotation}, we compare the performance of FV variants and {\it OxfordNet} {\it fc6} as queries are rotated at different angles.
Note that FVDoG is robust to rotation - the minor modulation in performance is due to filtering artifacts in the DoG interest point detector. 
However, the {\it OxfordNet} features, FVDS and FVDM have a steep drop in performance as queries are rotated.
The {\it OxfordNet} features are more rotation invariant than FVDS and FVDM.
Finally, note the large gap in performance between FVDoG and other schemes for the {\it Graphics} data set.
The gap in performance on {\it Graphics} arises from two contributing factors: (a) the worse performance of the {\it OxfordNet} features on this data set, and (b) the fact that there are several rotated queries on which the {\it OxfordNet} features perform worse.
The effect of each can be isolated from the next set of experiments we conduct.
Next, we discuss how to gain invariance for the CNN, FVDS and FVDM pipelines
Ideally, we desire invariance to rotation, while maintaining high discriminability.

\textbf{How do we gain rotation invariance for CNN features?}

\begin{figure}[ht]
\centering{
	\begin{tabular}{@{}c@{}@{}c@{}}
		\includegraphics[width=0.49\textwidth]{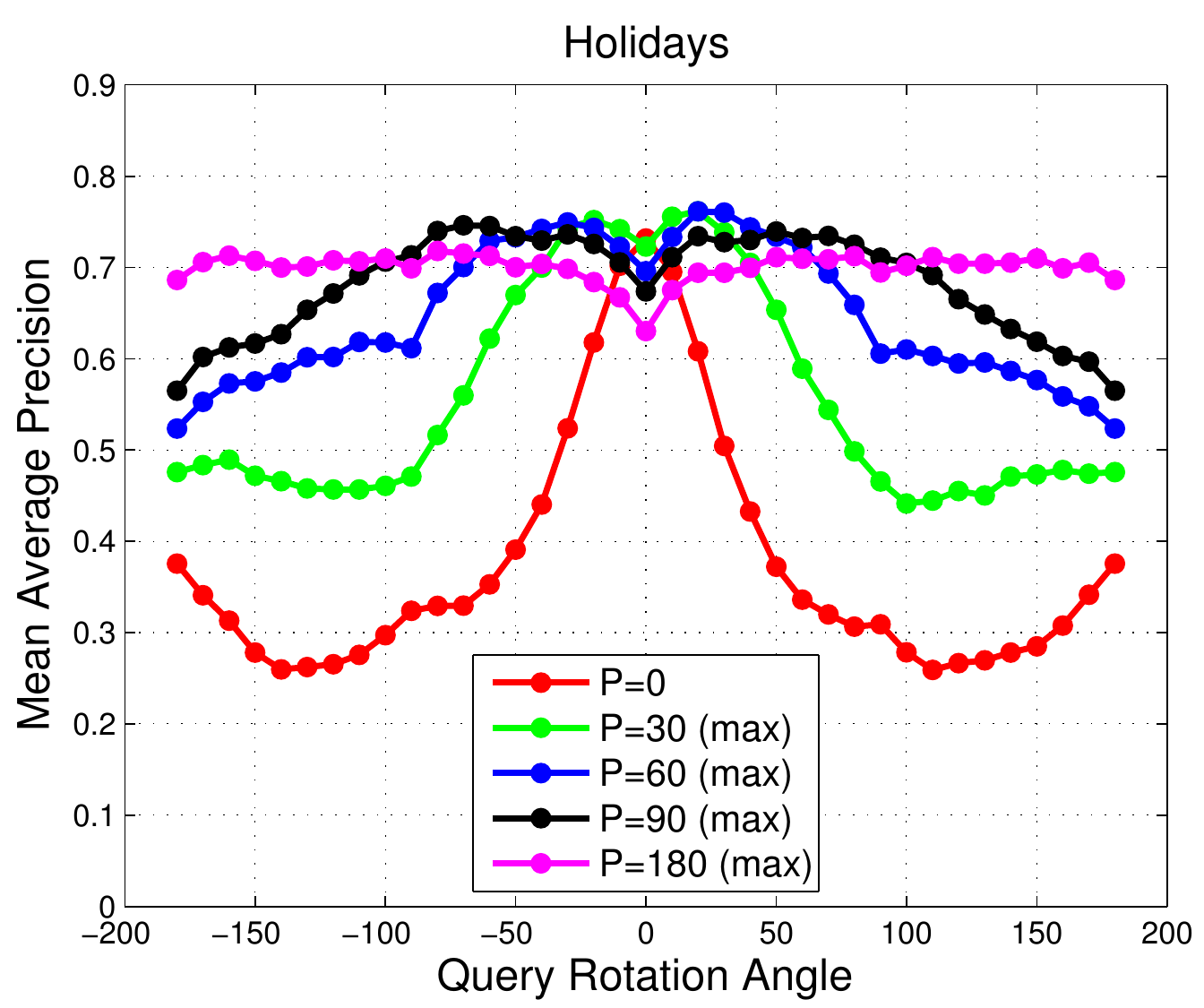} &
		\includegraphics[width=0.49\textwidth]{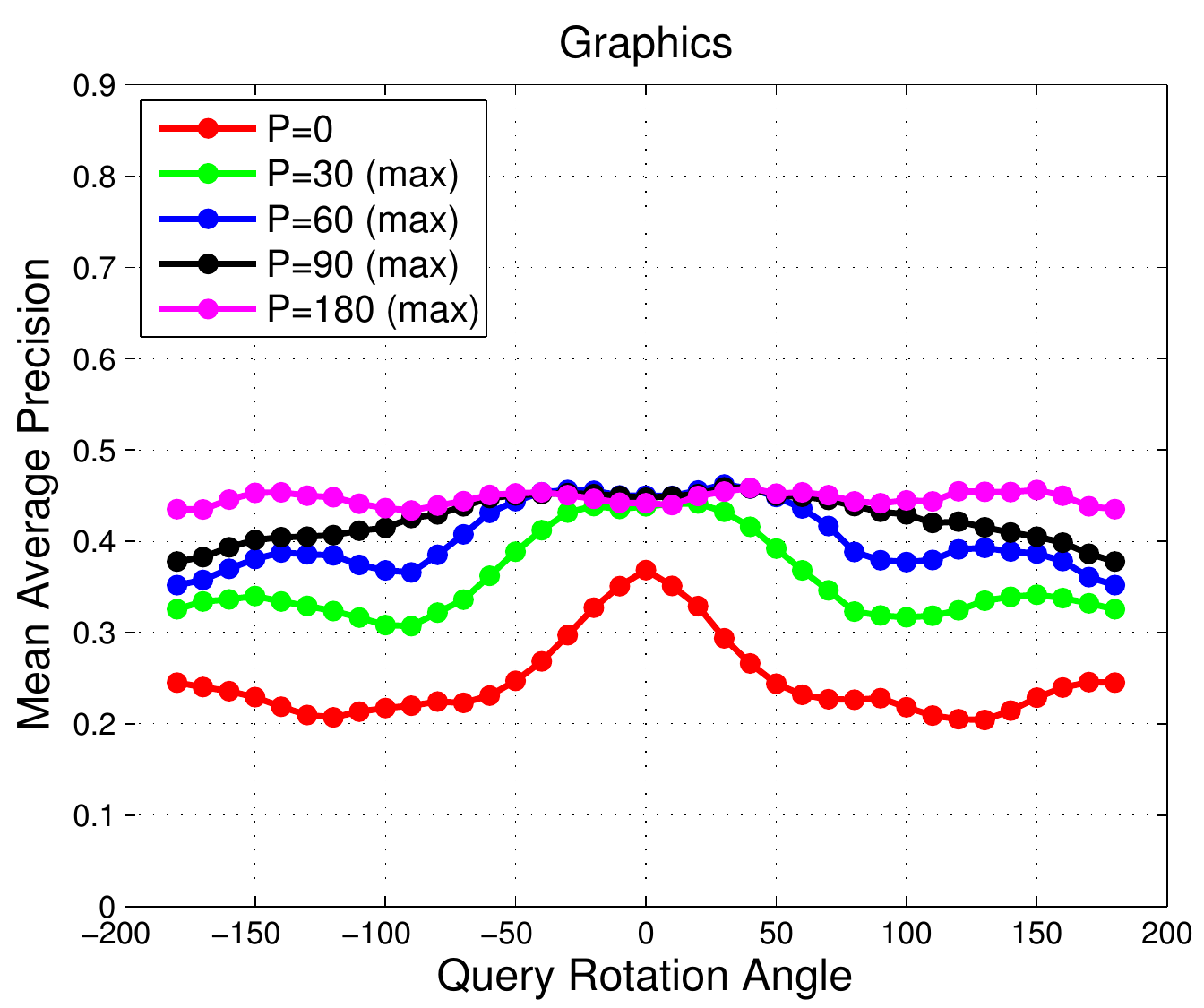} 
	\end{tabular}
	\caption{	
	MAP vs query rotation angle for different pooling parameters. Results are presented on {\it OxfordNet}  $fc6$ layer on the {\it Holidays} data set. $P=0$ refers to no pooling. $P=p$ refers to max-pooling over individual feature dimensions, for rotations between $-p\degree$ and $p\degree$ in steps of $s=10\degree$.  Invariance to rotation increases with increasing $p$, at the expense of lower performance at angle 0$\degree$.
		}
\label{fig:rot_pooling}
}	
\end{figure}

We propose a database pooling scheme illustrated in Figure~\ref{fig:cnn_rot_illus} for gaining rotation invariance.
Each database image is rotated within a range of $-p\degree$ to $p\degree$, in steps of 10$\degree$.
The CNN features for each rotated database image are pooled together into one common global descriptor representation.
In Figure~\ref{fig:rot_pooling}, we present results for max-pooling, where we store the component-wise maximum value across all rotated representations in an angular range.
$P=0$ refers to no pooling, while $P=p$ refers to pooling in the range of $-p\degree$ to $p\degree$ in steps of $s=10\degree$.
The parameter $s$ indicates the quantization step size of angular rotation of database images.

We plot performance as query rotation angle is varied for varying pooling parameter $P$, on {\it OxfordNet} {\it fc6} layer for the {\it Holidays} and {\it Graphics} data sets.
The invariance-discriminativeness trade-off is shown in Figure~\ref{fig:rot_pooling}.
We observe that the max pooling scheme performs surprisingly well for gaining rotation invariance.
As $P$ is increased, the performance curve flattens in the range of $-P\degree$ to $P\degree$, at the expense of lower performance for upright queries, i.e, at angle 0.
For the {\it Holidays} data set, most database and query images share similar ``upright'' orientations.
For the {\it Graphics} data set, note the gap in performance of different schemes at angle 0, between no pooling and different pooling schemes. 
This gap in performance can be attributed to rotated objects in the query data set.
The remaining gap at original angle 0$\degree$ between FVDoG in Figure~\ref{fig:cnn_vs_fisher_rotation} and CNN features in Figure~\ref{fig:rot_pooling} can be attributed to the worse performance of the {\it OxfordNet} features for this data set.

\begin{figure*}[ht]
\centering{
	\begin{tabular}{@{}c@{} @{}c@{}}
		\includegraphics[width=0.49\textwidth]{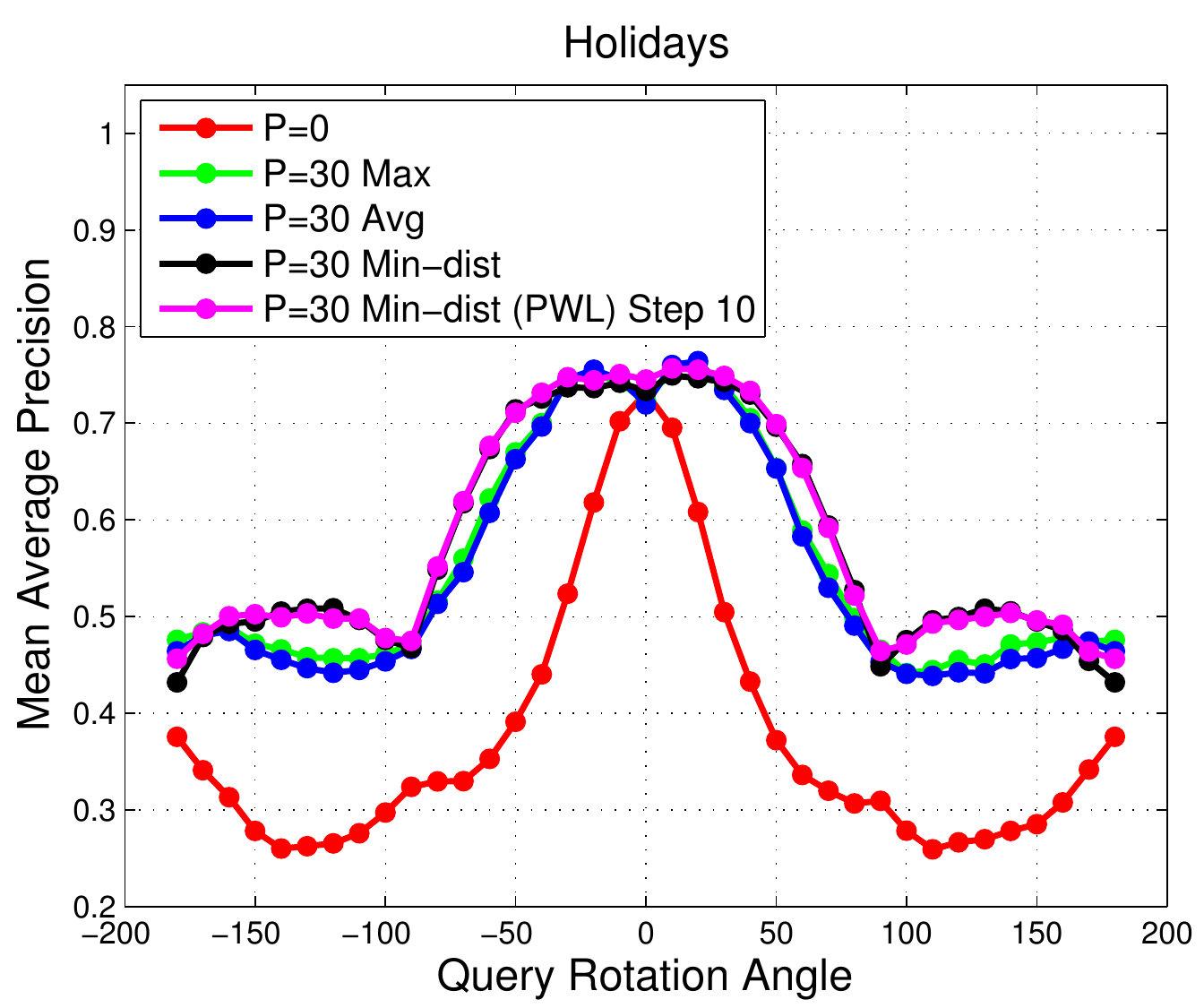} &
		\includegraphics[width=0.49\textwidth]{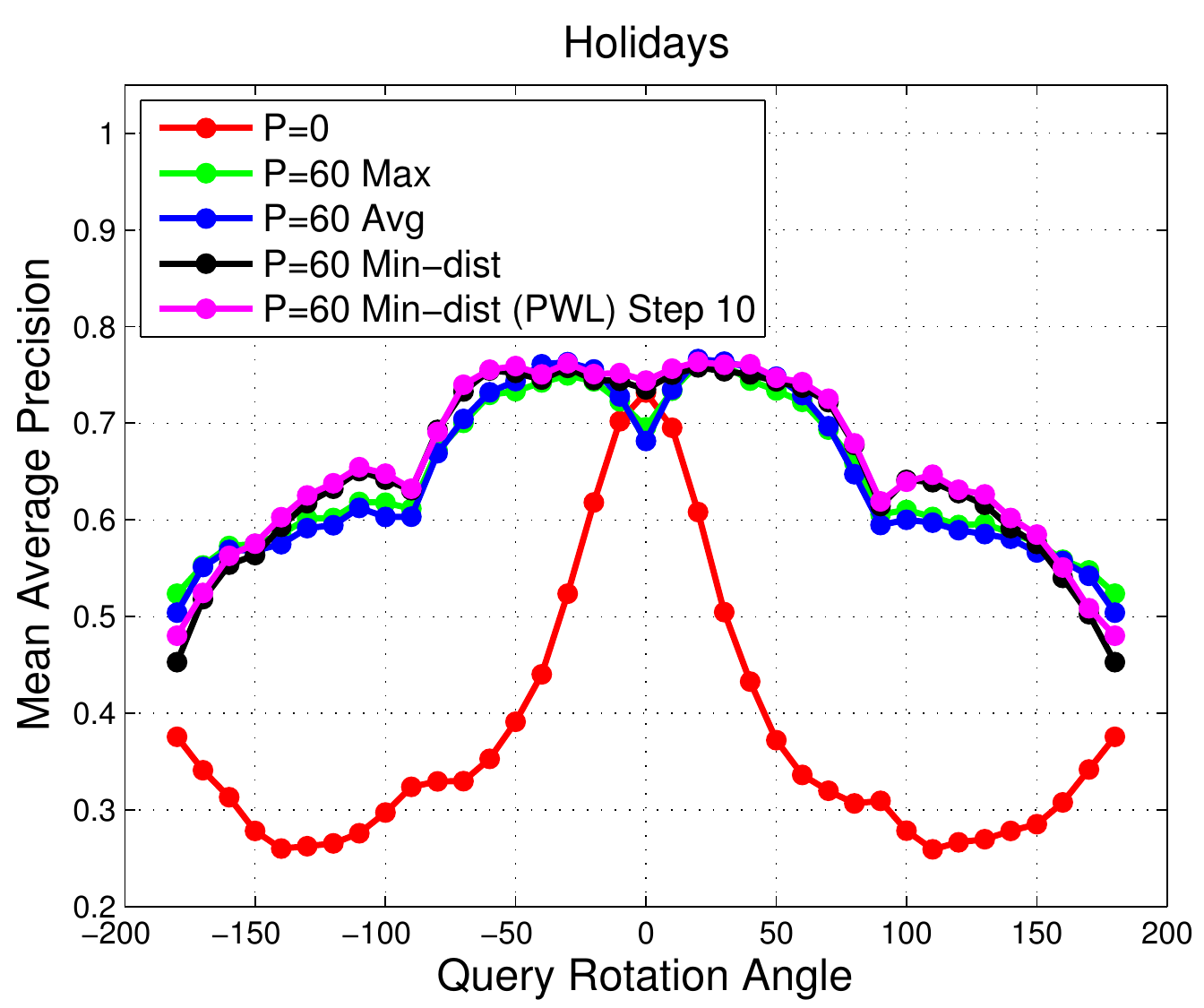} \\
		{\it $P=30\degree$} & {\it $P=60\degree$}\\
		\includegraphics[width=0.49\textwidth]{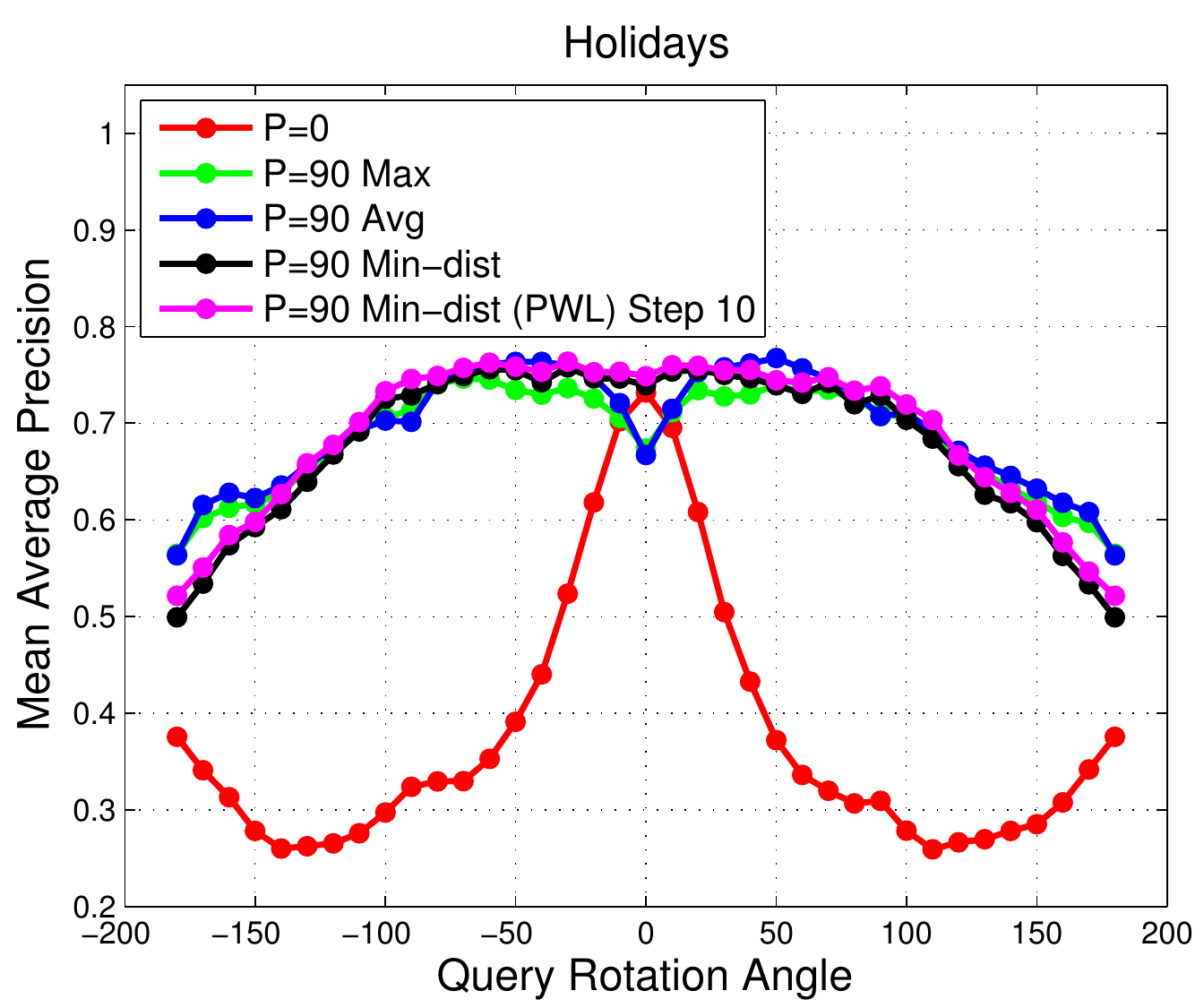} &
		\includegraphics[width=0.49\textwidth]{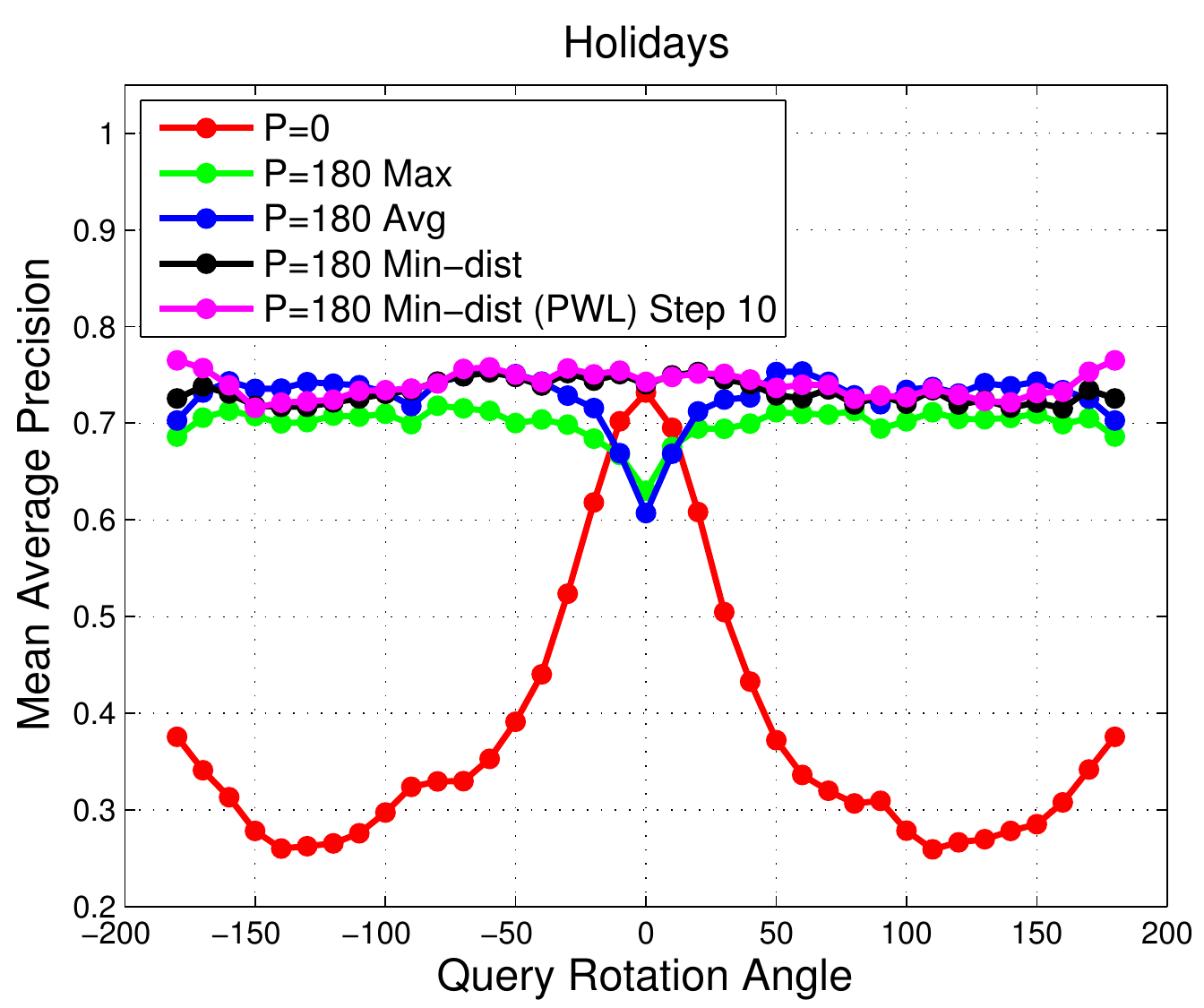} \\
		{\it $P=90\degree$} & {\it $P=180\degree$}
	\end{tabular}
	\caption{Comparison of different types of database pooling and augmentation techniques, for varying pooling parameter $P$. {\it OxfordNet} layer $fc6$ is used on the {\it Holidays} data set. We notice that max and average pooling come close to the perform of {\it Min-dist} and {\it Min-dist(PWL)}, which require storage of multiple feature descriptors ($\frac{2P}{s}+1$) for each database image, where $s=10$ is the quantization step size in degrees.
		}
\label{fig:cnn_rot_pooltype}
}	
\end{figure*}

To further understand the effectiveness of database-side pooling, we evaluate different types of pooling methods and database augmentation techniques in Figure~\ref{fig:cnn_rot_pooltype}.
We show results for component-wise max pooling and average pooling over rotated database images for different pooling parameters $P$.
We note that max pooling and average pooling perform comparably for small $P$. 
For $P=180\degree$, we note that average pooling outperforms max pooling.

\begin{figure}[ht]
	\centering{
		\includegraphics[width=0.49\textwidth]{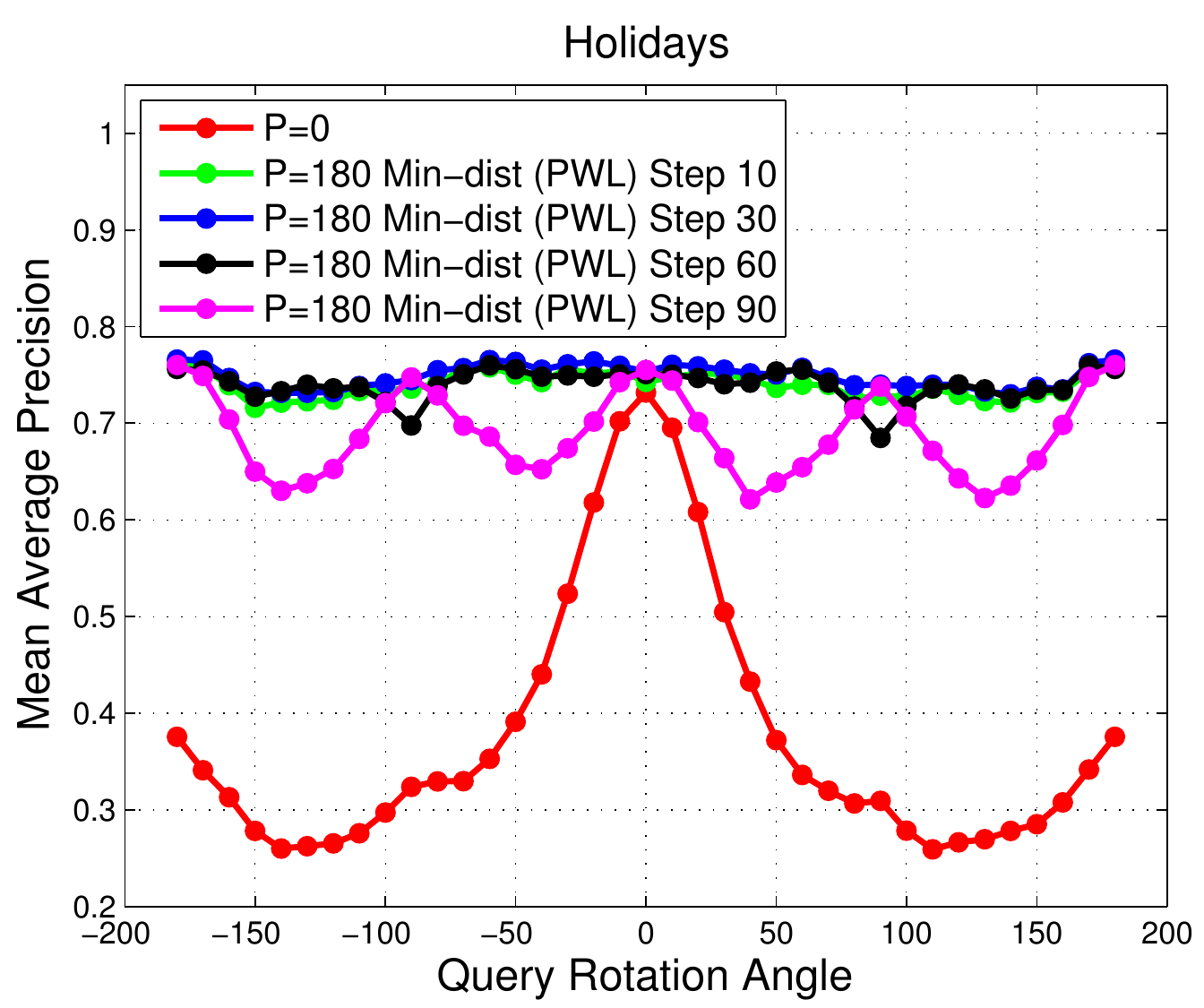}
		\caption{Results of the {\it Min-dist (PWL)} scheme as step size parameter $s$ is varied. {\it OxfordNet} layer $fc6$ is used on the {\it Holidays} data set. A piece-wise linear approximation of the manifold, on which rotated descriptors of each image lie, is used to trade-off performance and matching complexity. The performance of step size $s=60\degree$ is close to that of $s=10\degree$, while reducing memory requirements by 6$\times$.
	}
	\label{fig:pwl}
	}
\end{figure}

We compare the two pooling strategies to a simple database augmentation technique labeled {\it Min-dist}, which stores descriptors for each rotated version of the database image.
For {\it Min-dist}, at query time,  we compute the minimum distance to all the rotated versions for each database image.
The {\it Min-dist} increases the size of the database by $\frac{2P}{s}+1$, where $s=10$ is the step size in degrees, and $P$ is the pooling parameter.
For small $s \rightarrow 0$, the {\it Min-dist} scheme provides an approximate upper bound on the performance that the max and average pooling schemes can achieve, as a descriptor for each rotated version is explicitly stored in the database.
We observe that both max and average pooling are surprisingly effective, as their performance comes close to that of the {\it Min-dist} scheme while storing only one descriptor per database image.
Note that the {\it Min-dist} scheme performs the best, as there is no drop in performance at 0$\degree$, compared to the pooling methods.

Next, we also propose a scheme illustrated in Figure~\ref{fig:pwl_distance} for reducing memory requirements of the {\it Min-dist} scheme at the expense of increased matching complexity.
The scheme labeled {\it Min-dist (PWL)} assumes a piece-wise linear approximation of the manifold on which the descriptors of each rotated image lie, and computes the closest distance to the manifold.
The results for the {\it Min-dist (PWL)} with step size 10$\degree$ are shown in Figure~\ref{fig:cnn_rot_pooltype}, and it performs comparably to the {\it Min-dist} scheme.
Instead of maintaining database descriptors at finely quantized angular rotations of $s=10\degree$, we increase $s$ to $30\degree,60\degree,90\degree$ and present results in Figure~\ref{fig:pwl}.
We note that the performance of step size $s=60\degree$ is close to that of $s=10\degree$ for {\it Min-dist (PWL)}, while reducing memory requirements by 6$\times$.
The drop in performance for the {\it Min-dist (PWL)} scheme at -135$\degree$,-45$\degree$,45$\degree$,145$\degree$ for $s=90\degree$ shows inherent data set bias at these angles.
In conclusion, the proposed simple but elegant {\it Min-dist (PWL)} scheme helps gain rotation invariance, while requiring storage of fewer descriptors compared to the {\it Min-dist} approach.

\begin{figure}[ht]
	\centering{
		\includegraphics[width=0.49\textwidth]{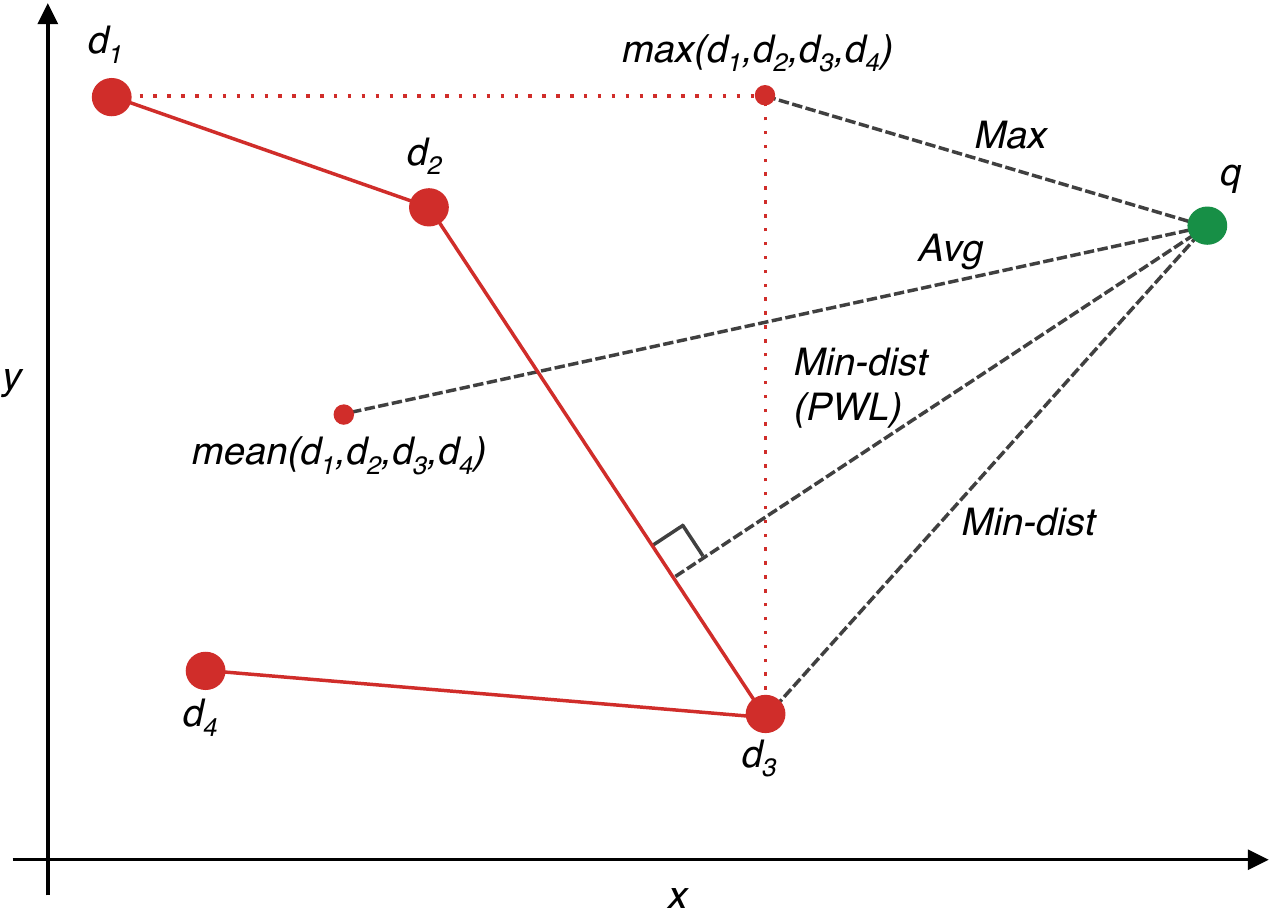}
		\caption{Illustration of the different pooling schemes. The scheme labeled {\it Min-dist (PWL)} assumes a piece-wise linear approximation of the manifold on which the descriptors of each rotated image lie, and computes the closest distance to the manifold.}
	\label{fig:pwl_distance}
	}	
\end{figure}

The surprising effectiveness of max and average pooling to gain rotation invariance for CNN features led us to run the same set of experiments on FVDM.
We present the results for max pooling, average pooling, {\it Min-dist}, and {\it Min-dist (PWL)} with step size $s=10\degree$ in Figure~\ref{fig:fv_pooltype} for $P=180\degree$.
Average pooling on FVDM helps gain invariance to rotation while lowering peak performance achieved without pooling ($P=0$).
However, note the large difference in performance between max and average pooling for FVDM.
CNN features are sparse with a small number of dimensions with high values: spikes resulting from the activation of neurons in the network.
FVDM data are comparatively more dense.
As a result, max pooling on {\it OxfordNet} features is far more effective than for FVDM.
Finally, in Figure~\ref{fig:fv_pooltype}, {\it Min-dist} and {\it Min-dist (PWL)} perform the best, as also observed for {\it OxfordNet} features.

\begin{figure}[ht]
	\centering{
		\includegraphics[width=0.49\textwidth]{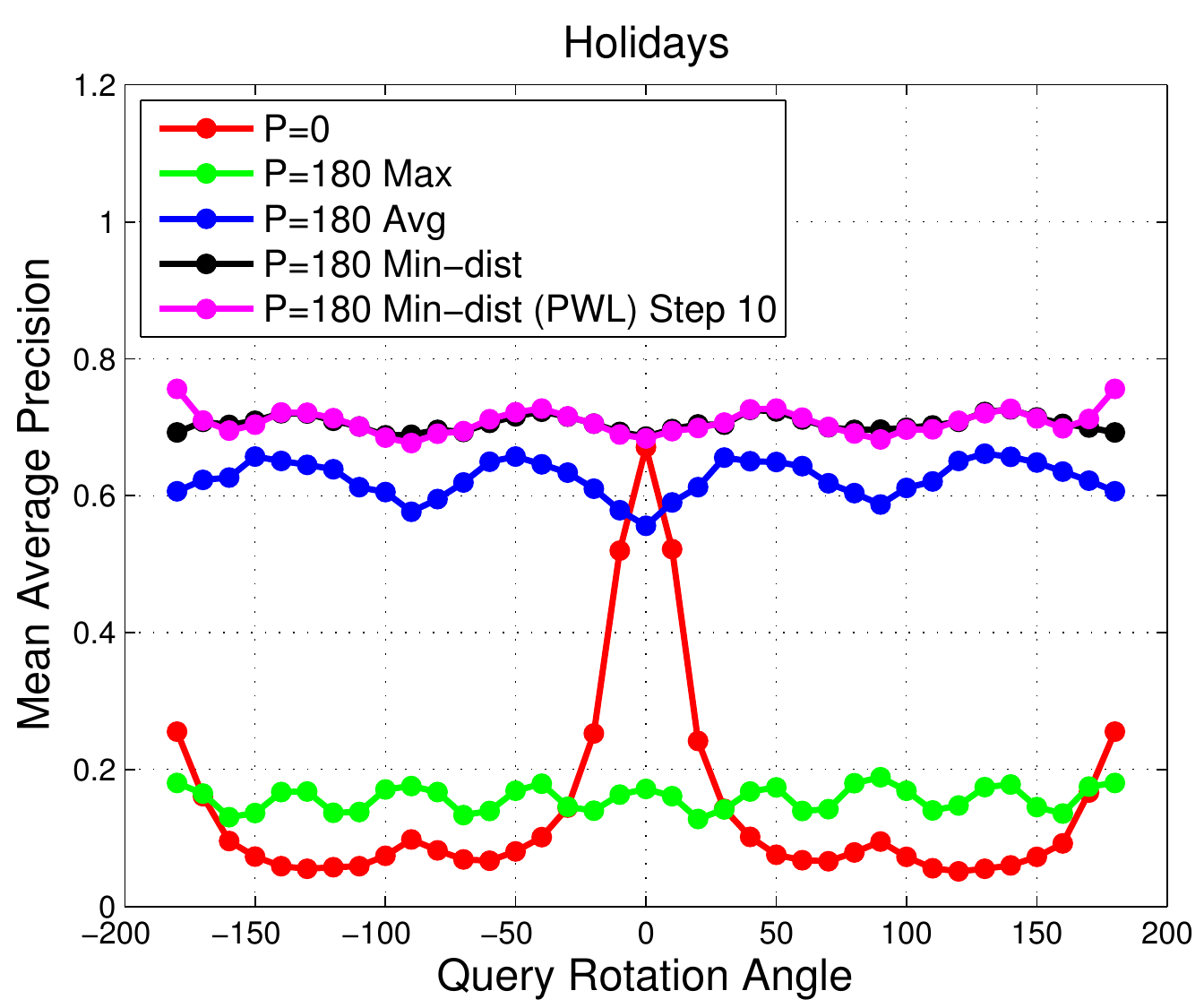}
		\caption{Comparison of different types of database pooling and augmentation techniques for pooling parameter $P=180$ for FVDM on the {\it Holidays} data set. Note the difference in performance of max and average pooling for FVDM, compared to max and average pooling on {\it OxfordNet} features in Figure~\ref{fig:cnn_rot_pooltype}(d). 
	}
	\label{fig:fv_pooltype}
	}
\end{figure}

\subsection{Invariance to Scale Changes}
\label{sec:scale-experiments}

In this section, we study scale invariance properties of CNNs and FVs.
Similar to rotation experiments in Section~\ref{sec:rotation_experiments}, we carry out control experiments on the {\it Holidays} data where we reduce the scale of query images and measure retrieval performance.
The starting resolution of all images (database and queries) is set to the larger dimension of 640 pixels (maintaining aspect ratio), as discussed in Section~\ref{sec:eval_framework}.

Both CNN and FV pipelines take in input images at fixed resolution.
For FVDoG, FVDS, FVDM pipelines, we resize images to VGA resolution (preserving aspect ratio) before feature extraction, even when the input resolution is smaller.
Upsampling images before feature extraction is shown to improve matching performance~\cite{Lowe04}.
The size of center crop input images (after resizing) to the {\it OxfordNet} pipeline is specified in Table~\ref{fig:cnns}.

\textbf{How invariant are CNN features to scale?} 

We scale query images along both image dimensions by a ratio of 0.75, 0.5, 0.375, 0.25, 0.2 and 0.125 starting from the VGA resolution - the smallest queries are $\left(\frac{1}{8}\right)^{th}$ the size of the VGA resolution image.
An anti-aliasing Gaussian filter is applied, followed by bicubic interpolation in the downsampling operation.
We present MAP for different layers of {\it OxfordNet} in Figure~\ref{fig:cnn_scale_invariance}.
We note that the $fc6$ layer experiences only a small drop in performance up to scale 0.25 before steeply dropping off.
The {\it OxfordNet} features are learnt on input images of fairly low resolution, which explains the robustness to large changes in scale.
We also note that the three fully connected layers exhibit similar characteristics for scale change: deeper fully connected layers are not more scale invariant.
It is interesting to note that $pool5$ is more scale invariant, as seen from the more gradual drop in performance as query scale is decreased: however, $pool5$ is less discriminative with a significant performance gap for smaller scale changes (0.75 to 0.25).

\begin{figure}[ht]
	\centering{
		\includegraphics[width=0.49\textwidth]{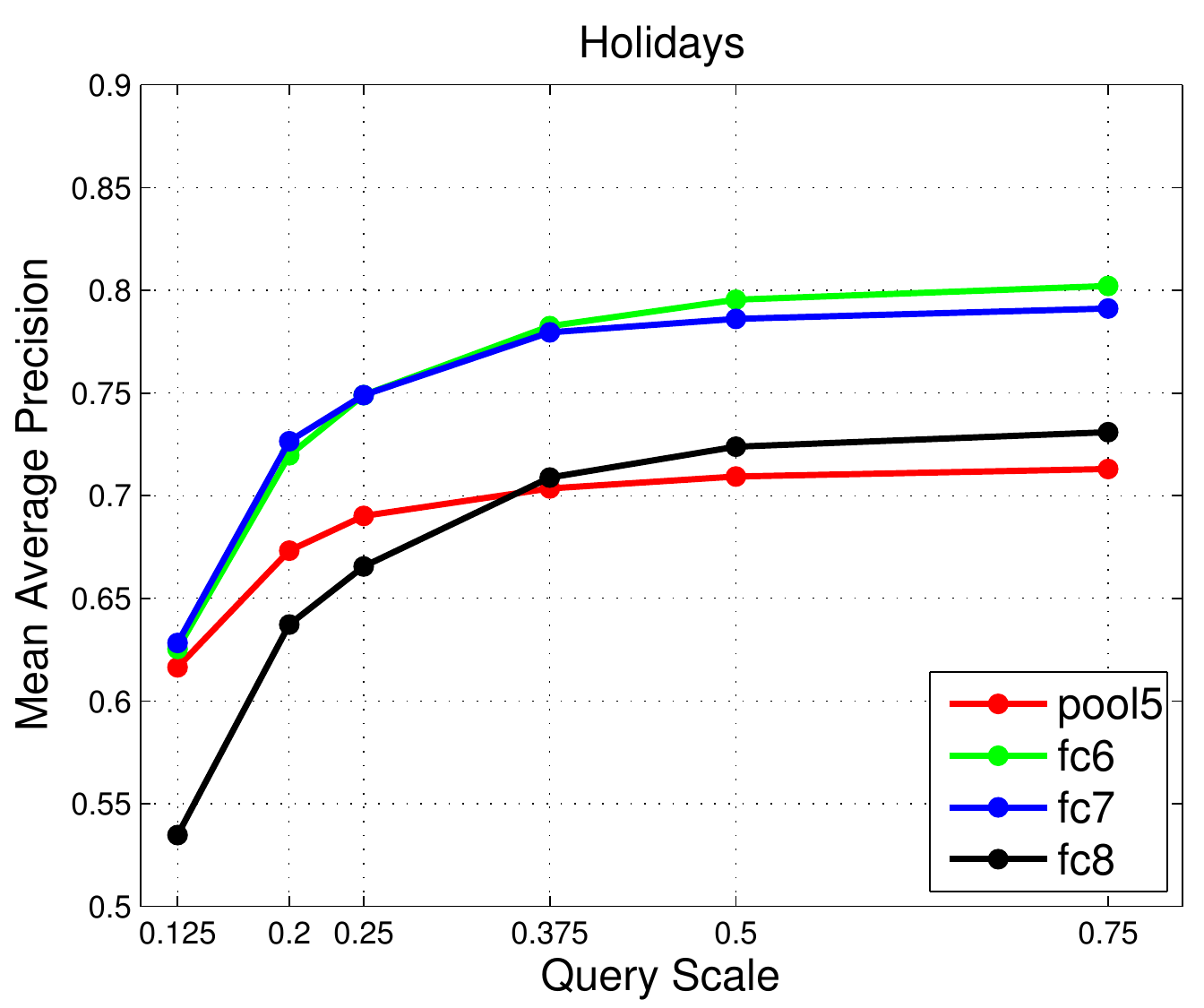}
		\caption{MAP as query images are scaled to 0.125 of original resolution, for different layers of {\it OxfordNet} on the {\it Holidays} data set. We note that {\it OxfordNet} features are robust to scale change up to 0.25, with performance dropping steeply after.
	}
	\label{fig:cnn_scale_invariance}
	}
\end{figure}

\textbf{Are CNN or FV more scale invariant?} 

Similar to the rotation experiment, we compare performance of {\it OxfordNet} $fc6$ and FVs in Figure~\ref{fig:cnn_vs_fv_scale} for the {\it Holidays} and {\it Graphics} data sets.
We observe a steeper drop in performance with decreasing scale for FVDM and FVDS compared to {\it OxfordNet}.
Somewhat surprisingly, FVDoG also experiences a sharper drop in performance compared to CNN.
The trends are consistent across data sets: the only difference is that the peak performance of FVDoG is higher than CNN on {\it Graphics}.
Trends similar to {\it Holidays} are observed on the remaining two data sets.
The sharp drop in performance of FVDoG can be attributed to the failure of the interest point detector at small scales.
CNNs are learnt on smaller images to begin with, and objects shown at different scales at training time, are sufficient for achieving more scale invariance than FVDoG.
In comparison to the rotation experiments, it is interesting to note that FVDoG are more robust to rotation changes, while {\it OxfordNet} features are more robust to scale changes.

\begin{figure}[ht]
\centering{
	\begin{tabular}{@{}c@{}@{}c@{}}
		\includegraphics[width=0.49\textwidth]{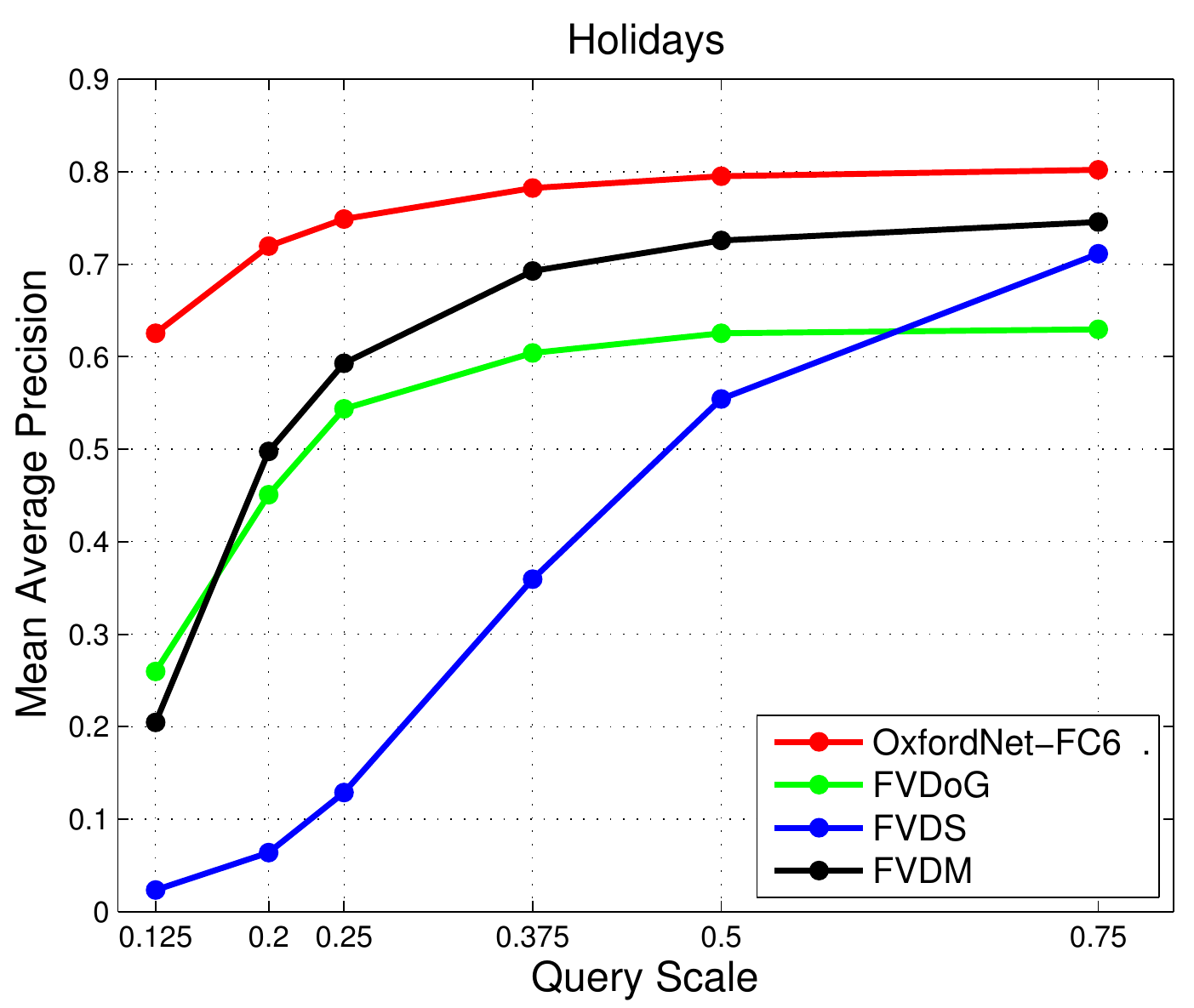}&
		\includegraphics[width=0.49\textwidth]{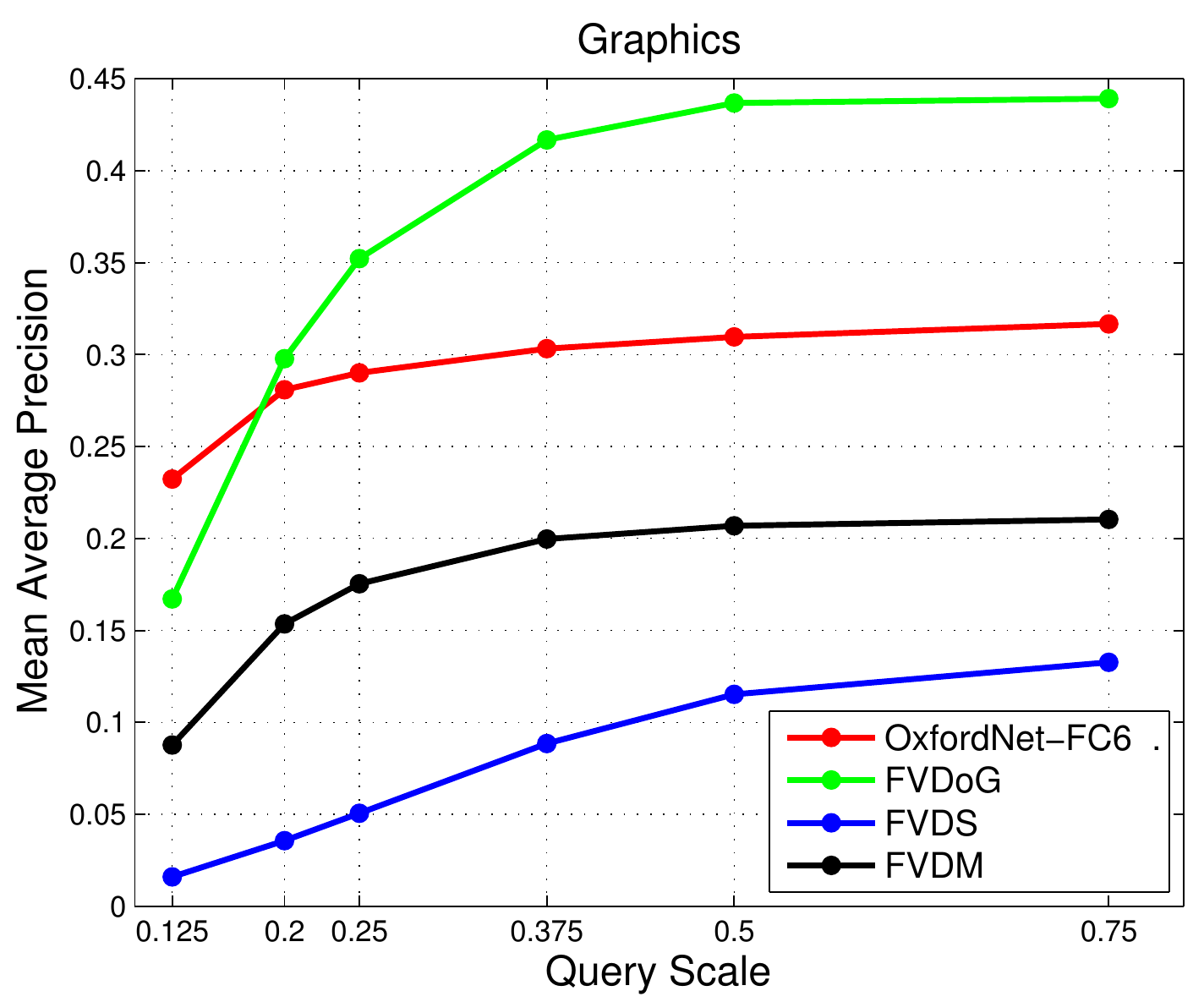}
	\end{tabular}
	\caption{Comparison of {\it OxfordNet} $fc6$ and FVs for scaled queries on the {\it Holidays} and {\it Graphics} data sets. We observe that {\it OxfordNet} features are more robust to scale changes compared to FVDoG, FVDS and FVDM, all of which experience a steeper drop in performance as query scale is decreased.
		}
\label{fig:cnn_vs_fv_scale}
}	
\end{figure}

\begin{figure}
	\centering{
		\includegraphics[width=0.7\textwidth]{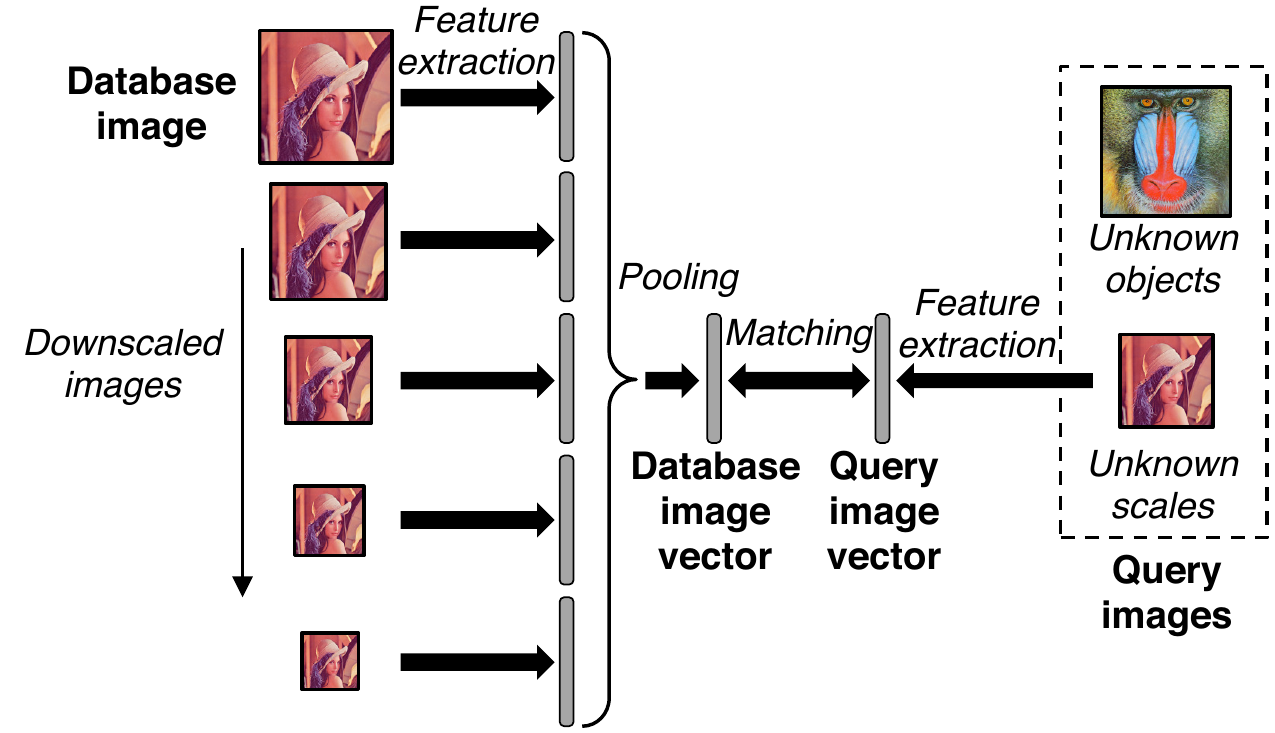}
		\caption{We extract features from database images at different scales and pool them to obtain a single representation. We scale queries to different sizes, and evaluate retrieval performance for different pooling parameters and strategies. 
	}
	\label{fig:cnn_scale_illus}
	}
\end{figure}

\textbf{Does database-side pooling improve scale invariance for CNN features ?}

Next, we discuss how performance at small scales can be improved by pooling descriptors on the database side.
As illustrated in Figure~\ref{fig:cnn_scale_illus}, the component-wise pooling operation across scales is similar to the database-pooling performed on rotated images.
The parameter $SP$ refers to the number of scales over which {\it OxfordNet} features are pooled.
$SP=n$ refer to pooling across the first $n+1$ scales of the set of six scale-ratios (seven including one) $\left(1,0.75,0.5,0.375,0.25,0.2,0.125)\right)$.
$SP=1$, hence, refers to no database pooling.

In Figure~\ref{fig:cnn_scales_pool}(a), we first study MAP vs query scale for different types of pooling on the {\it Holidays} data set for $SP=6$ (pooling over all scales in consideration).
{\it OxfordNet} {\it fc6} features are used in this experiment.
We note that max-pooling outperforms average pooling by a small margin, and comes close to the performance of the {\it Min-dist} scheme, which stores the descriptors of all the scaled versions of the database image and computes the minimum distance.
Similar to the rotation experiment, the {\it Min-dist (PWL)} scheme, which computes the minimum distance to a piece-wise linear manifold of the CNN descriptors for the six scaled images, is also effective for the scale experiment.
{\it Min-dist (PWL)} outperforms {\it Min-dist} by a small margin, as it is more robust to matching query data which lie at intermediate quantized scales.
For $SP=6$, there is a significant improvement in performance at small scales for the pooling schemes, with only a marginal drop in performance for points close to the original scale (seen from the right most points on the curve in Figure~\ref{fig:cnn_scales_pool}(a)).

In Figure~\ref{fig:cnn_scales_pool}(b), we study varying pooling parameter $SP$ for max-pooling.
Performance at small scales increases as $SP$ is increased, with only a marginal drop at query scale $0.75$.
A significant gain in performance of 10$\%$ is achieved for the smallest query scale $0.25$, showing the effectiveness of the max-pooling approach.

\begin{figure}
\centering{
	\begin{tabular}{@{}c@{} @{}c@{}}
		\includegraphics[width=0.49\textwidth]{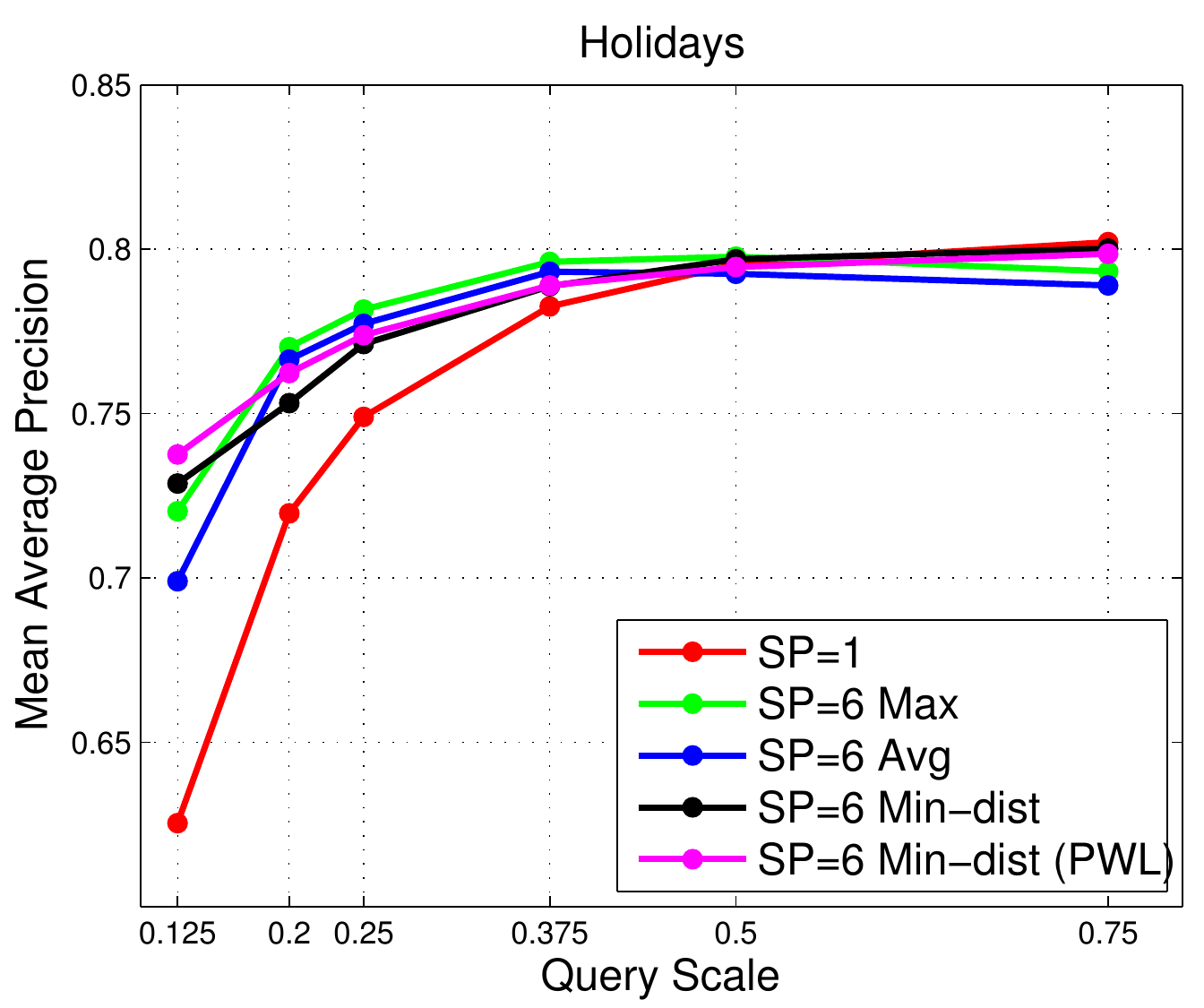} &
		\includegraphics[width=0.49\textwidth]{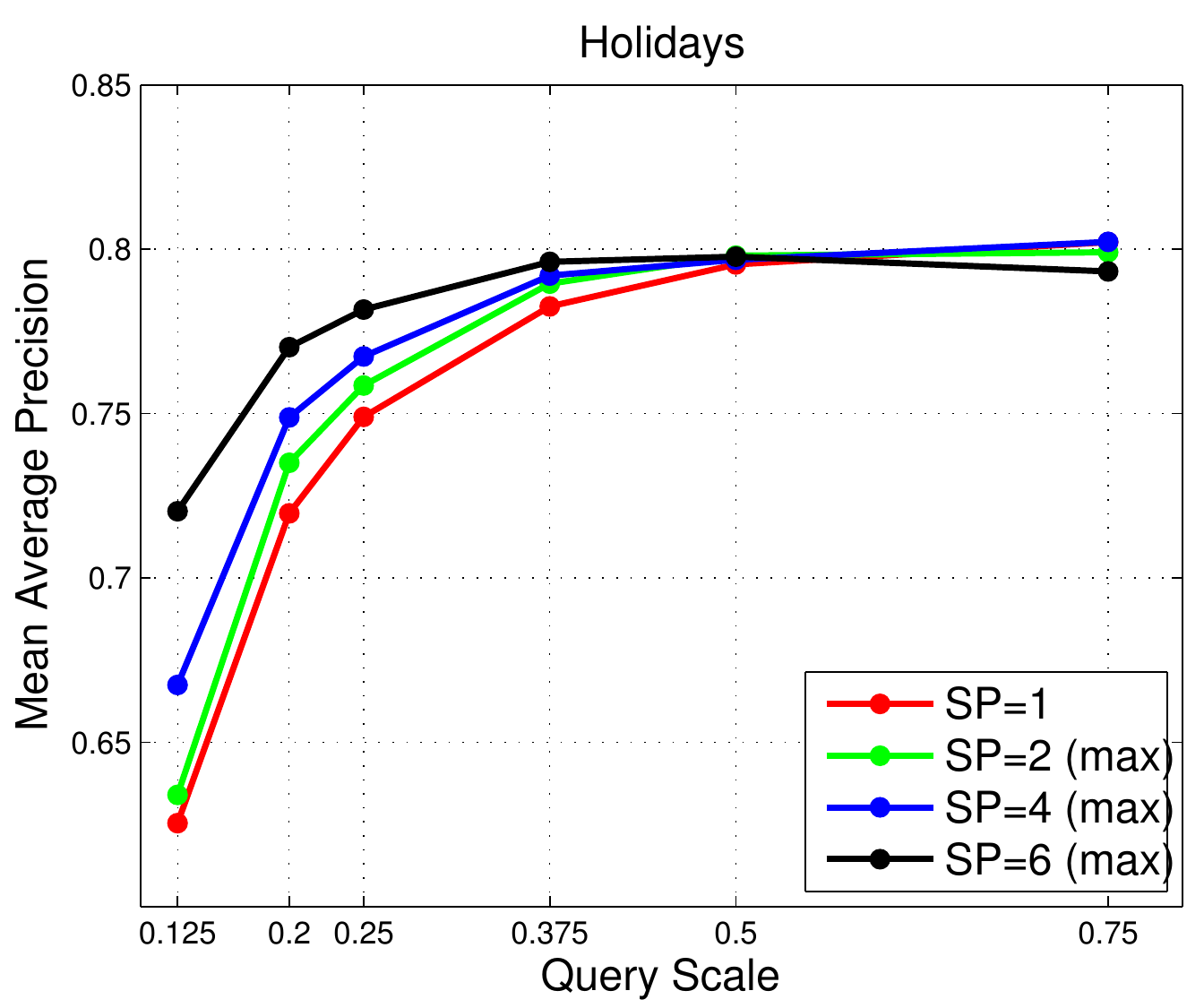} \\		
		(a) & (b) \\
	\end{tabular}
	\caption{Performance of different database-side pooling schemes, as query scale is changed. Results reported on {\it OxfordNet} $fc6$ on the {\it Holidays} data set. $SP=1$ refers to no database-pooling. $SP=n$ refer to pooling across the first $n+1$ scales of the set $\left(1,0.75,0.5,0.375,0.25,0.2,0.125)\right)$. In Figure (a), we notice that max pooling comes close to the performance of {\it Min-dist} which requires storing descriptors at all scales.  In Figure (b), we observe that performance improves at small scales with database side pooling as parameter $SP$ is increased.}
\label{fig:cnn_scales_pool}
}	
\end{figure}

\section{Open Questions}

The systematic study in this paper opens up several interesting avenues for future work.
We highlight the most important open questions here.

\begin{itemize}

\item Pre-trained CNN models trained for large-scale image classification tasks, with larger amounts of data have the potential of improving performance further for instance retrieval.
For instance, CNN models trained on the full ImageNet data set with 14 million images and 10000 classes could lead to more discriminative features for the instance retrieval task.

\item While supervised CNN models have far outperformed their unsupervised CNN counterparts for large-scale image classification, the latter approach deserves careful attention in the context of instance retrieval.
For the instance retrieval task, we desire rich representations of low level image information, which can be learnt directly from the large amounts of unlabelled image data available on the internet.
As image classification is not the end goal, unsupervised CNN models trained with large amounts of data might achieve comparable or better performance for instance retrieval tasks.
Availability of large amounts of training data (e.g., the Yahoo 100 million image data set~\cite{Yahoo100Mdataset}) and recent advances in open-source software for large-scale distributed deep learning (e.g. Torch~\cite{Torch}) will enable training of large-scale unsupervised CNN models.
If unsupervised CNN models work well for instance retrieval, they will enable easier training and adaptation to different types of image databases.

\item Rotation and scale invariance are key to instance retrieval tasks.
While the database pooling schemes proposed in this work are highly effective, they are more of an after-thought to solving the invariance problem in the CNN context.
Learning CNN representations which are inherently scale and rotation invariant is an exciting direction to pursue.

\item Interest point detectors provide an efficient and effective way of achieving desired levels of invariance (ranging from scale and rotation invariance to affine invariance).
The carefully hand crafted SIFT descriptor has been remarkably effective for the instance retrieval task: however patch level descriptors can now be learnt with large amounts of data, using data sets like the Winder and Brown patch data sets~\cite{WinderBrown09}, and the Stanford Mobile Visual Search patch data set~\cite{CDVSPatches}.
A hybrid approach of interest point detectors with learnt CNN descriptor representations could lead to a significant improvement in retrieval performance.

\item Hybrid interest point detection schemes like the dense interest point detector proposed originally in~\cite{DenseInterestPoints} need to be revisited, in light of the effectiveness of CNN features which are extracted by dense sampling in the image.
A recent survey of dense interest point detectors~\cite{DenseInterestPointsSurvey} is a good starting point.

\item Finally, our study has demonstrated that unlike for large scale image classification, combining CNNs with ``less effective'' types of descriptors such as FVs is a valid way to improve retrieval performance.
We point to readers the recent work from Gong et al.~\cite{NewPaperA} and Xu et al.~\cite{NewPaperB} in this field who have proposed effective FV/VLAD style encoding schemes for CNN descriptors.

\end{itemize}

\section{Conclusions}
\label{sec:conclusions}

In this work, we proposed a systematic and in-depth evaluation of FV and CNN pipelines for image retrieval.
Our study has lead to a comprehensive set of practical guidelines we believe can be useful to anyone seeking to implement state-of-the-art descriptors for image retrieval.
Some of the recommendations are general good practices while others are more problem specific.

We also showed that unlike image classification, the supremacy of CNNs over FVs does not always verify in the case of image retrieval and strategies mixing both approaches are most likely optimal.
In particular, the lack of transformation invariance of the descriptors appears to be one of the main drawbacks of CNNs.
We managed to propose a number of simple and effective approaches which can be followed to patch these deficiencies.
Nevertheless, we believe that better integrating invariance is key to the improvement of performance.

\bibliography{./main}   
\bibliographystyle{IEEEtran}

\end{document}